\pdfoutput=1
\documentclass[11pt,a4paper]{article}

\input{macros.sty}
\usepackage{authblk}
\usepackage[letterpaper, margin=1in]{geometry}
\setcitestyle{square,numbers}

\myexternaldocument{supplement}

\author[1]{Abhinav Gupta}
\author[1]{Pierre F.J. Lermusiaux}

\affil[1]{Department of Mechanical Engineering, 
Center for Computational Science and Engineering,
Massachusetts Institute of Technology, Cambridge, MA 02139\\
guptaa@mit.edu,\\ pierrel@mit.edu (Corresponding author)}

\date{\today}
\title{Generalized Neural Closure Models with Interpretability}

\begin{document}

\date{\today}
\maketitle

%%%% Abstract text to be placed here %%%%%%%%%%%%
\begin{abstract}
Improving the predictive capability and computational cost of dynamical models is often at the heart of augmenting computational physics with machine learning (ML). However, most learning results are limited in interpretability and generalization over different computational grid resolutions, initial and boundary conditions, domain geometries, and physical or problem-specific parameters. 
In the present study, we simultaneously address all these challenges by developing the novel and versatile methodology of unified neural partial delay differential equations. We augment existing/low-fidelity dynamical models directly in their partial differential equation (PDE) forms with both Markovian and non-Markovian neural network (NN) closure parameterizations. 
The melding of the existing models with NNs in the continuous spatiotemporal space followed by numerical discretization automatically allows for the desired generalizability. The Markovian term is designed to enable extraction of its analytical form and thus provides interpretability. The non-Markovian terms allow accounting for inherently missing time delays needed to represent the real world. 
Our flexible modeling framework provides full autonomy for the design of the unknown closure terms such as using any linear-, shallow-, or deep-NN architectures, selecting the span of the input function libraries, and using either or both Markovian and non-Markovian closure terms, all in accord with prior knowledge.
We obtain adjoint PDEs in the continuous form, thus enabling direct implementation across differentiable and non-differentiable computational physics codes, different ML frameworks, and treatment of nonuniformly-spaced spatiotemporal training data. 
We demonstrate 
%the abilities of 
the new \textit{generalized neural closure models} (\textit{g}nCMs) framework using four sets of experiments based on advecting nonlinear waves, shocks, and ocean acidification models. Our learned \textit{g}nCMs discover missing physics, find leading numerical error terms, discriminate among candidate functional forms in an interpretable fashion, achieve generalization, and compensate for the lack of complexity in simpler models. 
Finally, we analyze the computational advantages of our new framework.
\end{abstract}
%%%%%%%%%%%%%%%%%%%%%%%%%%%
%%%% Keyword entries to be placed here %%%%
%%%%%%%%%% Insert the texts which can accomdate on firstpage in the tag "fmtext" %%%%%

% \begin{fmtext}

%%% Put part of introduction here later on
%%%%%%%%%%%%%%%%%% Introduction %%%%%%%%%%%%%%%%%%%%%%%
\section{Introduction} \label{sec:intro}
%auto-ignore
%\PFJL{In this intro and abstract, please color all sentences that are the same or almost the same as in nCM paper 1. You can then provide ideas on what we should do with them, e.g.\ remove or rephrase, and we can discuss.}
%
%\AG{Using the following colors:} %\textcolor{cyan}{for exact ones,}\textcolor{orange}{for rephrased.}
%
%\PFJL{Ok, thanks, but perhaps this is more than what I was asking, but it may be useful.}

%\PFJL{We need to focus on what is in this paper, add motivating questions, etc. The first paragraph needs to introduce what is really in this paper, not complex and closure modeling alone. For examples of what is new (need better English): interpretable PDE terms, learning missing terms by separation of the ML into two parts, instantaneous and delayed (Markovian and non-Markovian that can be introduced a bit later to not have too much jargon in para 1), generic types of PDE applications considered (not many have unknown state variables), etc.}

%\PFJL{Major challenges addressed: A major challenge in SciML and closure modeling is to obtain analytical expressions, at least for some of the terms that are missing in the original PDEs. List other challenges. Some missing terms can be interpretable, some perhaps not, explain why. The instantaneous missing terms do not need MZ since they are Markovian.}

%\PFJL{We may need to organize the intro, i.e. what do we want to say where. I can help with all of the above, but I likely need to re-read the whole paper first. The slides may also help.}

The field of Scientific Machine Learning (SciML; \cite{SciMLwebsite}) is burgeoning with innovative methods that combine machine learning with existing scientifically-derived differential equation models and computational physics schemes. 
%to make realistic simulations of real-world phenomena feasible, given current computational capabilities. 
% 
This is in part because many realistic dynamical models are complex, and often truncated, coarsened, or aggregated due to computational cost constraints. Machine learning (ML) is then used to learn and represent the neglected and unresolved terms in a data-driven fashion \cite{gupta_lermusiaux_PRSA2021, pan2018data, pawar2020data, san2018neural, wan2018data, wang2020recurrent, sirignano2020dpm, saha2021deep}. Such techniques that express the missing dynamics as functions of modeled state variables and parameters are referred to as closure models.
Most ML closure models (and SciML results in general) are however often limited both in interpretability as black-box ML models and in generalization over different computational grid resolutions, initial conditions, boundary conditions, domain geometries, and physical or problem-specific parameters.
Addressing the challenges of interpretability and generalization is imperative
to justify the costs of training the SciML models using data sets obtained from expensive measurements or generated by solving the complex dynamical models in the first place.
The goal of the present study is to simultaneously address these challenges and learn closure models which are both generalizable and interpretable.

% Complex dynamical systems are used for predictions in many domains. However, due to computational
% cost constraints, models are often truncated, coarsened, or aggregated. As the neglected and unresolved
% terms become important, the utility of model predictions diminishes. There is a great deal of research on methods to model the missing dynamics and, with the advent of machine learning, there has been a renewed interest to learn the missing dynamics in a data-driven fashion \cite{pan2018data, pawar2020data, san2018neural, wan2018data, wang2020recurrent, sirignano2020dpm, saha2021deep}. Such techniques that express the missing dynamics as functions of modeled state variables and parameters are referred to as closure models.

The need for closure modeling 
%in dynamical systems
arises for a variety of reasons, ranging from computational cost considerations, preference for simpler models over complex ones due to overparameterization, or lack of scientific understanding of processes and variables involved in the system of interest. 
The simpler or the known model is often referred to as a low-fidelity model, while the complex counterpart in either models or observations is then referred to as the high-fidelity model, reality, or real-world data. %Encompassing various scenarios, 
Low-fidelity models can be categorized into three categories:
%\begin{enumerate*}
%  \item 
  (i) Reduced-order models, in which the original high-dimensional dynamical system is projected and solved in a reduced space. While it is computationally cheaper to solve the low-dimensional system, these models can quickly accumulate errors due to the missing interactions with the truncated dimensions \cite{feppon_lermusiaux_SIMAX2019,feppon_lermusiaux_SIREV2018, sapsis_lermusiaux_PHYSD2012};
  %\item 
  (ii) Coarse-resolution models, in which we only resolve the scales of interest. In these cases, the neglected and unresolved scales, along with their interactions with the resolved ones, can lead to unintended or unacceptable effects at global scales \cite{laizet2015influence, yeung2018effects, dauhajre2019nearshore, mcwilliams2016submesoscale, mcwilliams2017submesoscale};
  (iii) Simplistic or speculative models, in which an incomplete representation or understanding of processes and interactions occurs, and thus uncertainty in the model formulations and even in the relevant state variables themselves. This can lead to a gross or incorrect approximation of the real-world phenomena \cite{fennel2014introduction, may2019stability, nowak2006evolutionary,lermusiaux_et_al_O2006b,lermusiaux_PhysD2007}.
% \item Simplification of complex dynamical systems to reduce the complexity or number of state variables, components, and parameterizations \cite{fennel2014introduction, may2019stability, nowak2006evolutionary};
% \item Incomplete understanding leading to processes and interactions still hidden to scientists. 
%\end{enumerate*}

% \AG{This para contains the motivating question but they are not popping out and need rephrasing.}

In \cite{gupta_lermusiaux_PRSA2021}, neural closure models (nCMs) are developed for low-fidelity models using neural delay differential equations (nDDEs) and data from high-fidelity simulations. 
The need for time delays in closure parameterizations is rooted in the presence of inherent delays in real-world systems \cite{glass2020nonlinear, tokuda2019reducing} and theoretically justifed by the Mori-Zwanzig formulation \cite{chorin2000optimal, gouasmi2017priori, stinis2015renormalized, whitney1936differentiable}. 
Using nDDEs for closure modeling has a number of advantages. They allow for the use of smaller architectures and account for the accumulation of numerical time-stepping error in the presence of neural networks (NNs) during training. Additionally, nDDEs are agnostic to the time-integration scheme, handle unevenly-spaced training data, and have good performance over prediction periods much longer than the training or validation periods. 
However, there are other highly-desirable properties, as mentioned above. Fundamental questions for neural closures include: 
Can they be interpretable and lead to analytical expressions?
Can they achieve generalization over many conditions and
variables, as physics-based models do?
How can they be combined seamlessly with classic numerical schemes?
%
%different computational grid resolutions, initial conditions, boundary conditions, domain geometries, and physical or problem specific parameters.
% Moreover, when it comes to the melding of computational physics and machine learning (scientific machine learning; SciML \cite{SciMLwebsite}), there are other desirable properties as well, such as generalization over different computational grid resolutions, initial conditions, boundary conditions, domain geometries, physical or problem specific parameters, and interoperability. 
% 
A number of recent approaches have aimed to address such questions, however, challenges remain especially for partial differential equations (PDEs).
This is often because NNs are %typically 
used with the discretized ordinary differential equation (ODE) form of the corresponding PDEs, which makes it inherently difficult to generalize to changes in boundary conditions, domain geometry, and computational grid. Recently, a few studies have taken steps at addressing these drawbacks. 
Sirignano \textit{et al.} \cite{sirignano2020dpm} augment the underlying PDE with a neural network, however, they only learn a Markovian closure. The inputs to the neural network include the state, its spatial derivatives, and a fixed number of neighboring grid points. They also provide an accompanying discrete adjoint PDE for efficient training. Saha \textit{et al.} \cite{saha2021deep} use a radial-basis-functions-based collocation method to allow for mesh-free embedding of NNs. However, the resulting NNs also only learn a Markovian closure, do not account for the accumulation of time-integration errors, and lack interpretability.

% MB: Start this paragraph with "In this chapter, we propose ..."? 
In the present study, we develop the unified neural partial delay differential equations (nPDDEs) that augment existing/low-fidelity models in their PDE forms with both Markovian and non-Markovian closures parameterized with deep-NNs. 
The neural closure terms then contain instantaneous and delayed contributions. Their inputs consist of the modeled states, their spatial derivatives, combinations of derivatives, and any other problem-specific variables and parameters. 
The melding of the low-fidelity model and deep-NNs in the continuous spatiotemporal space automatically allows for generalizability to computational grid resolution, boundary conditions, and initial conditions. By design, the closure terms can also provide analytical expressions of the missing terms, thus leading to interpretability.
The resulting nPDDEs are discretized using 
%the desired numerical schemes.
%The results are directly applicable to 
any numerical method relevant to the dynamical system studied. 
Further, we provide adjoint PDE derivations in the continuous form, thus allowing one to implement across differentiable and non-differentiable computational physics codes, and also different machine learning frameworks. 
All our derivations and implementations are done considering deep-NN architectures, thus automatically encompassing linear- and shallow-NNs, and providing the user or subject-matter-expert user with the flexibility of choosing the architectural complexity in accord with the prior information available.
We refer to the new methodology as \textit{generalized} neural closure models (\textit{g}nCM).
Through a series of experiments, we demonstrate the flexibility of \textit{g}nCMs to learn closures either in an interpretable fashion, black-box fashion, or both simultaneously, using the prior scientific knowledge about the problem at hand. The \textit{g}nCMs can eliminate erroneous and redundant input terms, or combine them to achieve increased accuracy.  
We also demonstrate the generalizability of our learned closures to changes in physical parameters, grid resolution, initial conditions, and boundary conditions. 
Our first class of simulation experiments uses nonlinear waves and advecting shocks problems governed by the KdV-Burgers and classic Burgers PDEs. 
Our learned \textit{g}nCM  finds missing terms, discovers the leading truncation error, and a correction to the non-linear advection term. We find that training on data corresponding to just a few combinations of grid resolution and Reynolds number is sufficient to ensure that the learned closures are generalizable over a range of grid resolution and Reynolds number combinations, initial and boundary conditions, and also outperform the popular Smagorinsky subgrid-scale closure model. Our second class of experiments is based on ocean acidification models, where we learn the functional form of biological processes and compensate for the lack of complexity in simpler models obtained by aggregation of components and other simplifications of processes and parameterizations. Finally, we comment on the computational advantages of our new \textit{g}nCM framework.

In what follows, Section \ref{sec: Methodology} develops the theory and methodology for the gnCMs. Section \ref{sec: results and discussions} showcases the generalization and interpretability properties of the gnCMs in experiments with nonlinear waves,
advecting shocks, and ocean acidification, and discusses computational advantages. Conclusions are in Section \ref{sec: conclusions}.

%%%%%%%%%%%%%%%%%%%%%%%%%%%%%%%%%%%%%%%%%%%%%%%%%%%%%%%%%%%%%%%%%%%%%%%%%%%%%%%%%%%%%%%%%%%%%%%%%%%%%%%%%%%%%%%%%%%%%%%%%%%%%%%%%

%%%%%%%%%%%%%%%%%%%%%%%%%%%%%%%%%%%%%%%%%%%%%%%%%%%%%%%%%%%%%%%%%%%%%%%%%%%%%%%%%%%%%%%%%%%%%%%%%%%%%%%%%%%%%%%%%%
\section{Theory and Methodology}
\label{sec: Methodology}
%auto-ignore
The functional form of closure models representing missing dynamics can be derived by the Mori-Zwanzig formulation \cite{chorin2000optimal, gouasmi2017priori, stinis2015renormalized, whitney1936differentiable}, which proves it to be dependent on the time-lagged state dynamics. 
%Along with this formulation, many chemical or biological 
Many systems are modeled assuming smooth 
% concentration 
fields of state variables governed by % PDEs with fluid flow advection and/or mixing, leading to 
advection-diffusion-reaction PDEs.
% MB: The sentences above confused me. I tried to fix it based on what I thought you were saying, but just make sure I didn't write anything that is incorrect.
Such PDEs implicitly assume that local information between state variables is exchanged instantaneously at any spatial location. In reality, however, time delays occur for several reasons. First, reactions or changes in populations have non-negligible time scales. Such time delays are captured in more complex models by modeling intermediate state variables. The time response of lower-complexity models can thus approximate that of high-complexity models by explicitly introducing delays \cite{glass2020nonlinear, tokuda2019reducing}. 
Second, time delays arise due to missing subgrid-scale processes and/or truncated modes in reduced-order models. For all of these reasons, memory-based terms 
%are  needed to represent complex dynamical systems 
%The above arguments in favor of memory-based closure terms are thoroughly discussed in
and thus non-Markovian closure terms are needed to augment low-fidelity models
\citep{gupta_lermusiaux_PRSA2021}.

% The need for non-Markovian closure terms to augment low-fidelity models was motivated in the preceding paragraph. 
In general, low-fidelity models are also outright missing Markovian terms due to truncation, coarse resolution, or incomplete and uncertain functional forms of some of the model terms. 
We will therefore use both Markovian and non-Markovian terms to close low-fidelity models in their PDE forms. This leads to partial delay differential equations (PDDEs)
%; DDEs are a subclass of PDDEs, in the same fashion as ODEs are for PDEs). 
that are widely used in ecology, control theory, biology, and climate dynamics, to name a few application areas
%and are especially useful in situations where both spatial and temporal evolution matter 
\cite{wu2012theory}. 

In this study, the Markovian and non-Markovian closure terms will be modeled using deep-NNs. 
%MB: Tried to fix this sentences above but just check to make sure I didn't screw things up
% To achieve the interpretability from the learned weights of the NNs of the closure models, we consider closure terms that depend on the state variables, their spatial derivatives, and combinations of these belonging to a function library. 
%
To achieve full interpretability from the learned weights of the NNs of the closure models, we at times consider single-layer linear-NNs. Closure terms in general depend on the state variables, their spatial derivatives, and combinations of these belonging to a function library. 
As the presence of discrete delays can be seen as a special case of distributed delays,
the non-Markovian term is assumed to contain distributed delays and have a maximum finite time-delay ($\tau$). 
Given a continuous state vector comprising of $N_s$ different states, $u(x, t): \mathbb{R} \times [0, T] \rightarrow \mathbb{R}^{N_s}$, we thus consider a dynamical system belonging to domain $\Omega$ of the following form,
\begin{equation}
    \begin{split}
        \frac{\partial u(x, t)}{\partial t} &=
        \underbrace{\mathcal{L}\left(u(x, t), \frac{\partial u(x, t)}{\partial x}, \frac{\partial^2 u(x, t)}{\partial x^2}, ..., x, t; \nu \right)}_{Low-Fidelity~/~Known~Model} \\
        & + \underbrace{\mathcal{F}_{NN}\left(u(x, t), \frac{\partial u(x, t)}{\partial x}, \frac{\partial^2 u(x, t)}{\partial x^2}, ..., x, t; \phi \right)}_{Markovian~Closure~Term} \\
        & + \underbrace{\int_{t-\tau}^t \mathcal{D}_{NN}\left(u(x, s), \frac{\partial u(x, s)}{\partial x}, \frac{\partial^2 u(x, s)}{\partial x^2}, ..., x, s; \theta\right)ds}_{Non-Markovian~Closure~Term}, \quad x\in \Omega, \; t \geq 0\, , \\
        u(x, t) &= h(x, t), \; -\tau \leq t \leq 0 \quad \text{and} \quad \mathcal{B}(u(x, t)) = g(x, t), \; x \in \partial \Omega, \; t \geq 0 \,,
    \end{split}
    \label{eq: low-fid with closure terms}
\end{equation}
where $\mathcal{L}$, $\mathcal{F}_{NN}$, and $\mathcal{D}_{NN}$ are nonlinear functions parameterized with $\nu$, $\phi$, and $\theta$, respectively. $\nu$ are problem-specific parameters associated with the physical/biological/chemical phenomenon of interest, while $\phi$ and $\theta$ are the NN weights. 
When compared to PDEs, PDDEs require a history function ($h(x, t)\, , \; -\tau \leq t \leq 0 $) 
for their initialization at $t=0$. 
%
%\PFJL{(Add this ``$-\tau  \leq $'' everywhere in the IC equations? We don't need smaller t values.)}
%
%\AG{Yes, correct. Done}
%\PFJL{I found one missing above and added it, can you check again, search on leq?} 
%\AG{Done, thanks for catching it.}
%
The operator $\mathcal{B}$ represents appropriate boundary conditions such as Dirichlet, Neumann, etc. which are needed to solve the system uniquely. Furthermore, for ease of notation, we have assumed a one-dimensional (1D) domain, however, the method directly extends to 2D and 3D domains.

\subsection{Neural Partial Delay Differential Equations}
\label{sec: Neural Partial Delay Differential Equations}
We now obtain ML schemes that learn PDDEs parameterized using deep-NNs. They are referred to as \textit{neural} partial delay differential equations (\textit{n}PDDEs). Without loss of generality, and for brevity, we limit ourselves to \textit{n}PDDEs with only a Markovian term and a non-Markovian term with distributed delays. The low-fidelity model can be considered to be absorbed in the Markovian closure term. Hence, the \textit{n}PDDE is of the form,
\begin{equation}
    \begin{split}
        \frac{\partial u(x, t)}{\partial t} &= \mathcal{F}_{NN}\left(u(x, t), \frac{\partial u(x, t)}{\partial x}, \frac{\partial^2 u(x, t)}{\partial x^2}, ..., \frac{\partial^d u(x, t)}{\partial x^d}, x, t; \phi \right) \\
        & + \int_{t-\tau}^t \mathcal{D}_{NN}\left(u(x, s), \frac{\partial u(x, s)}{\partial x}, \frac{\partial^2 u(x, s)}{\partial x^2}, ..., \frac{\partial^d u(x, s)}{\partial x^d}, x, s; \theta\right)ds \,, \\
        & \hspace{0.6\textwidth} x\in \Omega, \; t \geq 0, \\
        u(x, t) &= h(x, t), \; -\tau \leq t \leq 0 \quad \text{and} \quad \mathcal{B}(u(x, t)) = g(x, t) \, \quad x \in \partial \Omega, \; t \geq 0 \;.
    \end{split}
    \label{eq: nPDDE in original form}
\end{equation}
%
%\PFJL{Instead of $\mathcal{G}_{NN}$, we could use $\mathcal{F}_{dNN}$ for delay-NN, and avoid GNN.}
%
%\AG{Or should we do $\mathcal{F}^d_{NN}$? Changed it to $\mathcal{D}_{NN}$}
%
The two deep-NNs, instantaneous 
$\mathcal{F}_{NN}(\bu; \phi)$ and delayed $\mathcal{D}_{NN}(\bu; \theta)$, remain
parameterized by $\phi$ and $\theta$, and for generality, they are considered to be functions of an arbitrary number of spatial derivatives, with the highest order defined by $d \in \mathbb{Z}^+$. 
We can rewrite equation \ref{eq: nPDDE in original form} as an equivalent system of coupled PDDEs with discrete delays,
\begin{equation}
    \begin{split}
        \frac{\partial u(x, t)}{\partial t} &= \mathcal{F}_{NN}\left(u(x, t), \frac{\partial u(x, t)}{\partial x}, \frac{\partial^2 u(x, t)}{\partial x^2}, ..., \frac{\partial^d u(x, t)}{\partial x^d}, x, t; \phi \right) + y(x, t) \;,  \\
        & \hspace{0.6\textwidth} x\in \Omega, \; t \geq 0 \,, \\
        \frac{\partial y(x, t)}{\partial t} &= \mathcal{D}_{NN}\left(u(x, t), \frac{\partial u(x, t)}{\partial x}, \frac{\partial^2 u(x, t)}{\partial x^2}, ..., \frac{\partial^d u(x, t)}{\partial x^d}, x, t; \theta\right) \\
        & - \mathcal{D}_{NN}\left(u(x, t-\tau), \frac{\partial u(x, t-\tau)}{\partial x}, \frac{\partial^2 u(x, t-\tau)}{\partial x^2}, ..., \frac{\partial^d u(x, t-\tau)}{\partial x^d}, x, t-\tau; \theta\right) \;,  \\
        & \hspace{0.6\textwidth} x\in \Omega, \; t \geq 0 \,,\\
        u(x, t) &= h(x, t), \; -\tau \leq t \leq 0 \quad \text{and} \quad\mathcal{B}(u(x, t)) = g(x, t), \; x \in \partial \Omega, \; t \geq 0 \,, \\
        y(x, 0) &= \int_{-\tau}^0 \mathcal{D}_{NN}\left(h(x, s), \frac{\partial h(x, s)}{\partial x}, \frac{\partial^2 h(x, s)}{\partial x^2}, ..., \frac{\partial^d h(x, s)}{\partial x^d}, x, s; \theta\right)ds \,.
    \end{split}
    \label{eq: nPDDE in the split form}
\end{equation}
Let us assume that high-fidelity data is available at $M$ discrete times, $T_1 < ...<T_M \leq T$, and at $N(T_i)$ spatial locations ($x_{k}^{T_i} \in \Omega, \forall k \in {1, ..., N(T_i)}$) for each of the times. Thus, we define the scalar loss function as, $L = \frac{1}{M} \sum_{i=1}^{M} \frac{1}{N(T_i)}\sum_{k=1}^{N(T_i)} l(u(x^{T_i}_k, T_i)) \equiv \int_0^T \frac{1}{M} \sum_{i=1}^{M} \int_{\Omega} \frac{1}{N(T_i)}\sum_{k=1}^{N(T_i)} l(u(x, t))\delta(x - x^{T_i}_k) \delta(t - T_i)dxdt \allowbreak \equiv \int_0^T \frac{1}{M} \sum_{i=1}^{M} \frac{1}{|\Omega|} \int_{\Omega} \hat{l}(u(x, t))\delta(t - T_i) dxdt$, where $l(\bu)$ are scalar loss functions such as mean-absolute-error (MAE), and $\delta(\bu)$ is the Kronecker delta function. 
% MB: I guess this is a minor detail but should there also be a second parameter in the loss function with the reference value (i.e. the high-fidelity data value).
In order to derive the adjoint PDEs, we start with the Lagrangian corresponding to the above system,
\begin{equation}
    \begin{split}
        \mathbb{L} &= L(u(x, t)) + \iw \lt(x, t) \left(\p_t u(x, t) - \mathcal{F}_{NN}(\bu, t; \phi) - y(x, t)\right) dxdt \\
        & + \iw \mt(x, t) \left(\p_t y(x, t) - \mathcal{D}_{NN}(\bu, t;\theta) + \mathcal{D}_{NN}(\bu, t-\tau; \theta)\right) dxdt \\
        & + \int_{\Omega} \alpha^T(x) \left(y(x, 0) - \int_{-\tau}^0 \mathcal{D}_{NN}(h(x, t), \p_x h(x, t), \p_{x^2} h(x, t), ..., \p_{x^d} h(x, t), x, t; \theta)dt\right)dx \,,
    \end{split}
\end{equation}
where $\lambda(x, t)$, $\mu(x, t)$ and $\alpha(x)$ are the Lagrangian variables. To find the gradients of $\mathbb{L}$ w.r.t. $\phi$ and $\theta$, we first solve the following adjoint PDEs (for brevity we denote, $\partial / \partial (\bu) \equiv \partial_{(\bu)}$, and $d / d (\bu) \equiv d_{(\bu)}$),
% MB: I think you mean in the brevity notation part to put the big dot not in the argument but as a subscript above?
\begin{equation}
\label{eq: gnCM adjoint PDEs}
    \begin{split}
        0 &= \frac{1}{M}\frac{1}{|\Omega|} \sum_{k=1}^M \p_{u(x, t)} \hat{l}(u(x, t)) \delta(t - T_k) \\
        & - \p_t \lt(x, t) -  \lt(x, t)\p_{u(x, t)}\mathcal{F}_{NN}(\bu, t) 
         + \sum_{i=1}^d (-1)^{i+1} \p_{x^i} \left(\lt(x, t) \p_{\p_{x^i} u(x, t)} \mathcal{F}_{NN}(\bu, t) \right)
        \\
        & - \mt(x, t) \p_{u(x, t)} \mathcal{D}_{NN}(\bu, t; \theta)
        + \sum_{i=1}^d (-1)^{i+1} \p_{x^i} \left(\mt(x, t) \p_{\p_{x^i} u(x, t)} \mathcal{D}_{NN}(\bu, t; \theta) \right)
         \\
        & + \mt(x, t+\tau) \p_{u(x, t)} \mathcal{D}_{NN}(\bu, t; \theta)
        - \sum_{i=1}^d (-1)^{i+1} \p_{x^i} \left(\mt(x, t+\tau) \p_{\p_{x^i} u(x, t)} \mathcal{D}_{NN}(\bu, t; \theta) \right) \;, \\
        & \hspace{0.6\textwidth} x \in \Omega \;, \; t \in [0, T) \,,
         \\
        0 & = -\lt(x, t) - \p_t \mt (x, t) \;, \quad x \in \Omega \;, \; t \in [0, T) \,, 
    \end{split}
\end{equation}
with initial conditions, $\lambda(x, t) =\mu (x, t) = 0, \; t \geq T$. The boundary conditions are derived based on those of the forward PDDE and they satisfy,
\begin{equation}
\label{eq: gnCM adjoint PDEs BCs}
    \begin{split}
        0 &= \sum_{i = 0}^d \sum_{j=0}^{d-i-1} (-1)^{j+1} \p_{x^{j}} \left(\lt(x, t)\p_{\p_{x^{j+i+1}} u(x, t)} \mathcal{F}_{NN}(\bu, t)\right)\dt \p_{x^{i}}u(x, t) 
        \\
        & +\sum_{i = 0}^d\sum_{j=0}^{d-i-1} (-1)^{j+1} \p_{x^{j}} \left(\mt(x, t)\p_{\p_{x^{j+i+1}} u(x, t)} \mathcal{D}_{NN}(\bu, t)\right)\dt \p_{x^{i}}u(x, t) \\
        & - \sum_{i = 0}^d\sum_{j=0}^{d-i-1} (-1)^{j+1} \p_{x^{j}} \left(\mt(x, t+\tau)\p_{\p_{x^{j+i+1}} u(x, t)} \mathcal{D}_{NN}(\bu, t)\right)\dt \p_{x^{i}}u(x, t) \;, \\
        & \hspace{0.6\textwidth} x \in \partial \Omega, \; t \in [t, T) \,.
    \end{split}
\end{equation}
Details of the derivation of the above adjoint PDEs are in section \ref{sec: gnCM Adjoint Equations in supp info}. After solving for the Lagrangian variables, $\lambda(x, t)$ and $\mu(x, t)$, we compute the required gradients as,
\begin{equation}
    \begin{split}
        \dt\mathcal{L} &= - \iw \mt(x, t) \pth \mathcal{D}_{NN}(\bu, t; \theta) dxdt + \iw \mt(x, t) \pth \mathcal{D}_{NN}(\bu, t-\tau; \theta) dxdt \\
        & - \int_{\Omega} \mt(x, 0) \int_{-\tau}^0 \pth \mathcal{D}_{NN}(h(x, t), \p_x h(x, t), \p_{xx} h(x, t), x, t; \theta)dt dx \,, \\
        d_{\phi}\mathcal{L} &= - \iw \lt(x, t) \pp \mathcal{F}_{NN}(\bu, t; \phi) dxdt \,.
    \end{split}
\end{equation}
Finally, using a stochastic gradient descent algorithm, we find the optimal values of the weights $\phi$ and $\theta$.

\subsection{Generalized Neural Closure Models: Properties}

\begin{figure}[h!]
  \centering
  \includegraphics[width=1\textwidth]{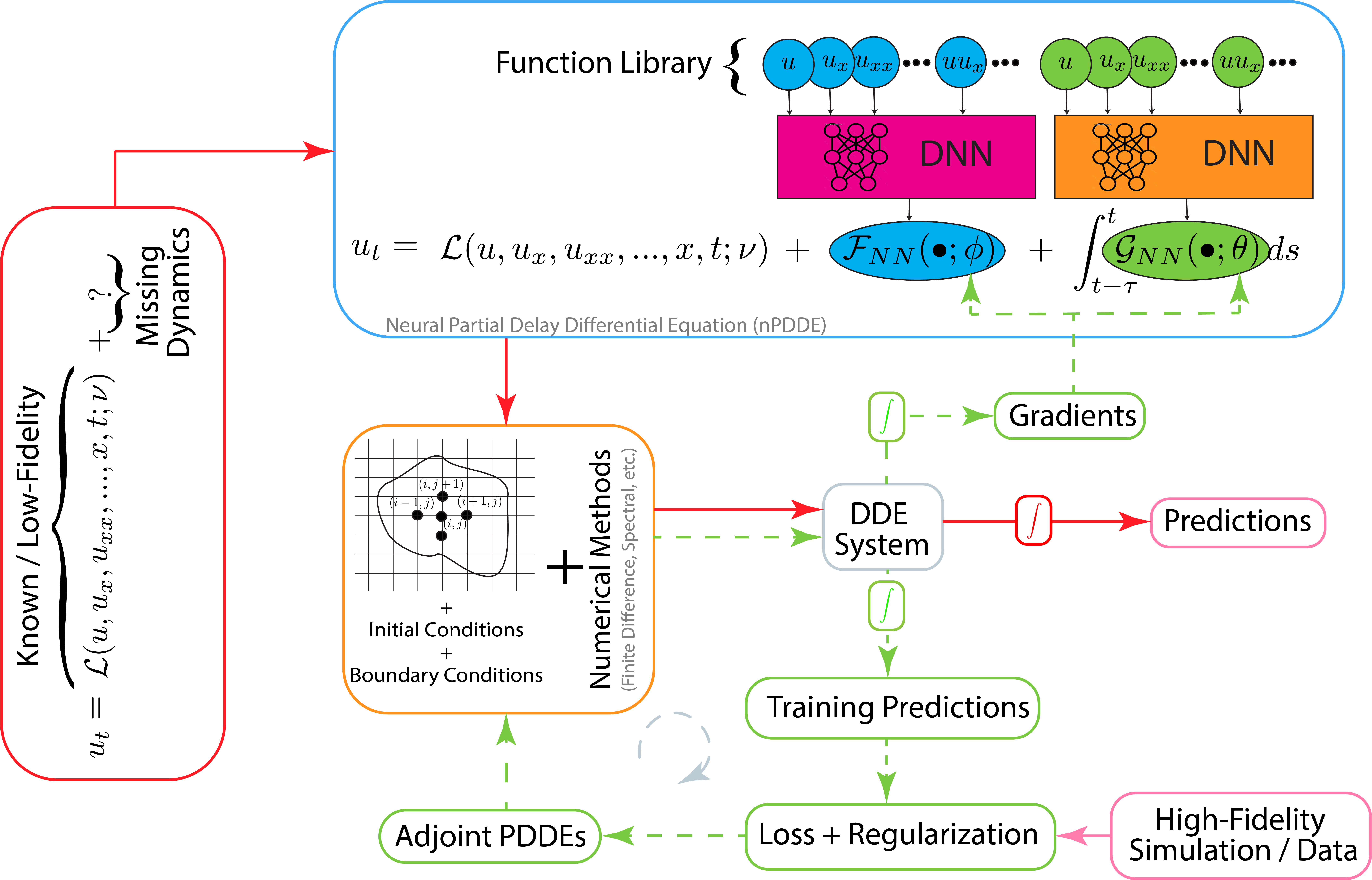}
  \caption{Overview of the \textit{generalized} neural closure models (\textit{g}nCM) framework. The blocks labeled \textit{DNN} represent any deep-neural-network architectures. The block labeled $\int$ symbolizes any time-integration scheme. DDE stands for delay differential equation.
  }
  \label{fig: framework overview}
\end{figure}

The \textit{g}nCM framework is schematized in Figure \ref{fig: framework overview}. Next, we discuss some of its properties and variations.

\paragraph{\textit{Interpretability.}} 
For interpretability -- especially for the Markovian closure term -- we can use a simple NN architecture with no hidden layers and linear activation. Nonlinearity can still be introduced by having input features that are nonlinear combinations of the states and their derivatives belonging to a function library. 
%\PFJL{Do we need to clarify this sentence so it fits better with the prior one?} 
%\AG{Added ``nonlinear" in the next line. Hope that makes it clear.} 
The result is a linear combination of these nonlinear input features. 
%
% Along with this, a $L_1$ regularization on the NN weights and pruning below a threshold helps promote sparsity.
% %
% The \textit{g}nCM method 
% allows for redundancy in the input function library and can update this input library adaptively \cite{kulkarni_et_al_DDDAS2020}.  In practice, one can include as many input test functions as computationally efficient and scientifically meaningful, and then adaptively augment and prune this library during the data-driven learning process.  
%
Along with this, a $L_1$ regularization on the NN weights and pruning below a threshold helps promote sparsity, thus allowing for redundancy in the input function library. In practice, one can include as many input test functions as computationally efficient and scientifically meaningful, and then adaptively augment and prune this library during the data-driven learning process, similar to that demonstrated in \cite{kulkarni_et_al_DDDAS2020}.
%
% Although this approach may seem similar to SINDy \cite{brunton2016discovering, rudy2019data, kulkarni_et_al_DDDAS2020}, it is different because it accounts for the accumulation of time-integration errors during training and does not require training data to be rich enough to allow for the computation of temporal and spatial derivatives.
%
Although this approach has similarities to SINDy \cite{brunton2016discovering, rudy2019data, kulkarni_et_al_DDDAS2020}, it is significantly different. 
SINDy requires training data to be rich enough to allow for the computation of temporal and spatial derivatives, and solves a regression problem to discover the governing dynamical system. 
Some successors of SINDy circumvent the need for calculating spatio-temporal derivatives from training data by utilizing weak forms \cite{messenger2021weak} and NNs to map coordinates of the problem to the state variable \cite{both2021deepmod}.
%
% The \textit{g}nCM method however accounts for the accumulation of time-integration errors during training by numerically solving the PDE augmented with the Markovian closure term and its corresponding adjoint PDE. It does not require using the training data to compute any temporal and spatial derivatives. 
Our \textit{g}nCM method also does not require using the training data to compute any temporal and spatial derivatives. It further accounts for the accumulation of time-integration errors during training by numerically solving the PDE augmented with the Markovian closure term and its corresponding adjoint PDE.
%The \textit{g}nCM 
Compared to other model discovery methods, \textit{g}nCM seamlessly incorporates and simultaneously learns a non-Markovian closure term without simplifying assumptions.

The use of an informative function library along with a simple NN architecture with no hidden layers and linear activation is also applicable for the non-Markovian term for enhanced interpretability.
In fact, in our derivation above (section~\ref{sec: Neural Partial Delay Differential Equations}) and framework implementation, we keep the possibility of using any general deep-NN architectures for both Markovian and non-Markovian closure terms. 
This allows one to introduce an arbitrary amount of nonlinearity, especially in cases when no prior information is available about the functional form of the missing dynamics. 
The use of deep-NNs comes at the cost of full interpretability. However, even in this case, some insight can be obtained, for example, by examining the weights of the input layer of the learned deep-NN to determine the relative importance of different input features. 
This is showcased in Experiments-1b (section~\ref{sec: Experiments 1b}) for the learned non-Markovian closure term.
%
%\PFJL{We may need to be clearer in how we differentiate our work w.r.t. SINDY.}
%\AG{Tried to make it clearer.}

\paragraph{\textit{Generalizability.}}
The forward model (equation \ref{eq: low-fid with closure terms} or \ref{eq: nPDDE in original form}) and the adjoint PDEs (equation \ref{eq: gnCM adjoint PDEs}) are discretized and integrated using numerical schemes \cite{chapra2011numerical}, such as finite differences, finite volumes, collocation methods, etc.
This new approach, where we augment the PDEs with the NN-based 
Markovian and non-Markovian closures first, before numerical discretization, ensures that the burden of generalization over boundary conditions, domain geometry, and computational grid resolution, along with computing the relevant spatial derivatives is handled by the numerical schemes, and not by the learned deep-NNs. 
This also automatically makes the learning only dependent on local features and affine equivariant, similar to numerical schemes. 
%\PFJL{In the macros, We can add your latex spacing control for the figure and captions so as to reduce the big white spaces, I didn't find them.} \AG{Added the commands in ms.tex}

\paragraph{\textit{Backpropagation and Adjoint Equations.}} 
With the \textit{adjoint method}, the adjoint PDEs (equations~\ref{eq: gnCM adjoint PDEs} \& \ref{eq: gnCM adjoint PDEs BCs}) are solved backward in time, and one would require access to $u(x, t), \forall x \in \Omega, \; 0 \leq t \leq T$. 
{In the original neural ODEs \cite{chen2018neural}, the proposed adjoint method forgets the forward-time trajectory $u(x, t), \forall x \in \Omega, \; 0 \leq t \leq T$; instead, it remembers only the state at the final time, $u(x, T)$, and then solves for $u(x, t)$ in reverse-time along with the adjoint PDEs.
% In the original neural ODEs \cite{chen2018neural}, the forward model was solved backward in time starting from the knowledge of $u(T)$, which has been shown to be a major source of errors. 
% The adjoint method forgets the forward-time trajectory $u(x, t), \forall x \in \Omega, \; 0 \leq t \leq T$; instead, it remembers only the state at the final time, $u(x, T)$, and then solves for $u(x, t)$ in reverse-time. 
This approach is known to suffer from inaccuracies and numerical instabilities \cite{gholami2019anode, zhuang2020adaptive}.
Thus, in our current implementation, we create and continuously update an interpolation function using the $u$ obtained at every time step as we solve the forward model (equation \ref{eq: nPDDE in original form}). 
For memory efficiency, one could also use the method of \textit{checkpointing} \cite{griewank1992achieving, gholami2019anode, zhuang2020adaptive}, or the interpolated reverse dynamic method (IRDM) \cite{daulbaev2020interpolation}. 
Along with this, using adaptive time-integration schemes leads to stable and accurate solutions for our forward and adjoint PDEs, especially for stiff dynamical systems \cite{zhang2022memory, enriquez2011effects}. 
The inherent inaccuracies and instabilities of using continuous adjoint equations followed by discretization remain open questions \cite{gholami2019anode, zhang2022memory, enriquez2011effects, zhuang2020adaptive, li2004adjoint} and well-known issues in data assimilation \cite{robinson_et_al_Sea1998,robinson_lermusiaux_Sea2002}. 
In this work, we found that the combination of the continuous adjoint method followed by discretization and adaptive time-integration schemes is successful. 
Another challenge that can occur is the feasibility of derivation of the continuous adjoint PDEs followed by discretization, especially for known realistic (low-fidelity) models that are highly complex and nonlinear. In such cases, the discrete adjoint method, i.e., the approach of deriving the adjoint equations for the discrete forward model might be more useful
\cite{lermusiaux_et_al_O2006a,moore_et_al_FMS2019}.
This makes it easier to utilize the vast array of tools developed by the \textit{Automatic Differentiation} community over the last several decades \cite{autodiff}, specifically, the source-code-transformation (source-to-source) methods \cite{van2017tangent, van2018automatic}.
Finally, reduced-space adjoints as well as ensemble approaches can be used to estimate gradients \cite{geer2021learning}. 

%%%%%%%%%%%%%%%%%%%%%%%%%%%%%%%%%%%%%%%%%%%%%%%%%%%%%%%%%
%%%%%%%%%%%%%%%%%%%%%%%%%%%%%%%%%%%%%%%%%%%%%%%%%%%%%%%%%

%%%%%%%%%%%%%%%%%%%%%%%%%%%%%%%%%%%%%%%%%%%%%%%%%%%%%%%%%
%%%%%%%%%%%%%%%%%%%%%%%%%%%%%%%%%%%%%%%%%%%%%%%%%%%%%%%%%

\section{Application Results and Discussion}
\label{sec: results and discussions}
%auto-ignore
Using four sets of experiments, we now showcase and evaluate the capabilities of our new closure modeling framework (\textit{g}nCM) in terms of generalizability over grid resolutions, boundary and initial conditions, and problem-specific parameters. We also demonstrate the 
interpretability of the learned closures within PDEs. 

In the first and second sets of experiments, we consider problems based on advecting nonlinear wave and shock PDEs. We find that \textit{g}nCMs can discriminate and discover processes such as dispersion, the leading truncation error term, and a correction to the nonlinear advection term, all in an interpretable fashion with the learned Markovian neural closure. Using both Markovian and non-Markovian neural closure terms, we demonstrate the generalization of the \textit{g}nCM over grid resolution, Reynolds number, and initial and boundary conditions, along with superior performance compared to the popular Smagorinsky closure. 
In the third and fourth sets of experiments, we consider problems based on coupled physical-biological-carbonate PDEs used to study the threat of ocean acidification. 
We utilize the \textit{g}nCMs to discriminate and discover the functional form of uncertain terms with interpretability, and to augment a simpler model obtained by aggregation of components and simplifications of processes and parameterizations, such that, it becomes as accurate as a more complex model. 

Our training and evaluation protocol is similar to that in \cite{gupta_lermusiaux_PRSA2021}. 
%In all experiments, 
%the training data is regularly sampled in both space and time from the high-fidelity simulations, however, this is not a requirement. 
We use performance over the validation period (past the period for which high-fidelity data snapshots are used for training) to fine-tune various training-related hyperparameters. The final evaluation is based on continuous evolution through the training and validation periods, followed by longer-term future predictions. 
% MB: Should it be "through" in the line above
We also compare the learned closure with the known true model.
%, whenever available. 
%In the rest of the paper, 
For all the figure, table, and section references prefixed with ``SI-'', we direct the reader to the \textit{Supplementary Information}.

\subsection{Experiments 1a: Nonlinear Waves - Interpretable Model Discrimination}
%auto-ignore
%\textcolor{blue}{Summ.:}
In the first set of experiments, we consider advecting nonlinear wave PDEs and discover missing/uncertain physical processes, such as dispersion, in an interpretable fashion, using the learned Markovian neural closure. 

\textcolor{black}{\emph{Setup: True model, data generation, and low-fidelity model.}} 
Models for advecting shocks and nonlinear waves are the backbone of various physical phenomena.
%such as surface waves. 
The Korteweg de Vries (KdV)-Burgers PDE is often used to describe weak effects of dispersion, dissipation, and non-linearity in such wave propagation \cite{shi2015numerical}. %In the first set of experiments
Here, considering a 1D spatial domain, we select this KdV-Burgers PDE as the high-fidelity model (truth),
\begin{equation}
\label{eq: true kdv equation}
    \begin{split}
        \frac{\partial u}{\partial t} = -6u\frac{\partial u}{\partial x} - \frac{\partial^3 u}{\partial x^3} \;.
    \end{split}
\end{equation}
The data is generated from two solitary waves colliding with each other and that are exact solutions of equation \ref{eq: true kdv equation} 
with initial and boundary conditions given by,
\begin{equation}
\label{eq: true kdv ic and bc}
    \begin{split}
        u(x, 0) &= 2 \eta_1^2 \text{sech}[\eta_1(x - x_1)] + 2 \eta_2^2 \text{sech}[\eta_2(x - x_2)], \\
        u(-L, t) &= 0, \; \frac{\partial u (x, t)}{\partial x} \bigg|_{x = L} = 0, \; \text{and} \; \frac{\partial^2 u (x, t)}{\partial x^2} \bigg|_{x = L} = 0 \,,
    \end{split}
\end{equation}
where $x_1$ is the location, $2 \eta_1^2 $ is the amplitude, and $1/\eta_1$ is the width of the first soliton wave, whereas $x_2$ is the location, $2 \eta_2^2 $ is the amplitude, and $1/\eta_2$ is the width of the second soliton wave, initially. 
The parametric analytical solution of the above system is given by,
\begin{equation}
\label{eq: kdv equation analytical solution}
    \begin{split}
        u(x, t) = \frac{8(\eta_1^2 - \eta_2^2) (\eta_1^2 \cosh{\theta_2} + \eta_2^2 \sinh{\theta_1})}{( (\eta_1 - \eta_2) \cosh(\theta_1 + \theta_2) + (\eta_1 + \eta_2) \cosh(\theta_1 - \theta_2))^2} \,,
    \end{split}
\end{equation}
where $\eta_1 \geq \eta_2$, and $\theta_1$ and $\theta_2$ are given by,
\begin{equation}
    \begin{split}
        \theta_1 &= \eta_1(x-x_1 - 4 \eta_1^2 t) \;, \\
        \theta_2 &= \eta_1(x-x_2 - 4 \eta_2^2 t) \;.
    \end{split}
\end{equation}
We choose $L=10$, maximum time $T = 1.5$, $\eta_1 = 1.2$, $\eta_2 = 0.8$, $x_1 = -6.0$ and $x_2 = -2.0$.

For the closure learning experiments, we assume we
only have prior knowledge about the existence of the advection term and the low-fidelity model is thus, 
\begin{equation}
\label{eq: just advection term}
    \begin{split}
        \frac{\partial u}{\partial t} = -u\frac{\partial u}{\partial x} \;.
    \end{split}
\end{equation}
Other effects are unknown and need to be discovered. We assume these unknown effects to be mainly Markovian in nature and that they can be modeled using a linear combination from a library of nonlinear functions comprising terms up to 3rd order spatial derivatives:
%and arising in the generalized KdV-Burgers PDE. 
%The library is thus:
$\left\{ \frac{\partial^2 u}{\partial x^2}, \frac{\partial^3 u}{\partial x^3}, u\frac{\partial u}{\partial x}, u^2 \frac{\partial u}{\partial x}\right\}$.
%The low-fidelity model is solved numerically which leads to truncation errors.
%
Compared to the true model (equation~\ref{eq: true kdv equation}), our library contains two superfluous or redundant terms,  $\frac{\partial^2 u}{\partial x^2}$ and $u^2 \frac{\partial u}{\partial x}$. Of course, it does not contain repetitive terms.

\textcolor{black}{\emph{Numerics.}}
All the numerical solutions of low-fidelity model augmented with the  \textit{g}nCM
are obtained using finite difference schemes. For the advection term, $2^{nd}$ order accurate upwind \cite{rahul2006one} is used, while all other spatial terms and derivatives are discretized with $4^{th}$ order accurate central-difference. For time-marching, the \textit{Vode} scheme \cite{brown1989vode} with adaptive time-stepping is used. Finally, we employ a fine grid with $N_x = 200$ number of grid points in the $x-$direction in order to keep low discretization and truncation errors. 
\textcolor{black}{\emph{Comparison LF-HF:}}
In Figure \ref{fig: Exp1a_analy_low-fid_diff},
we compare the numerical solution of the low-fidelity model (equation \ref{eq: just advection term}) with the analytical solution of the high-fidelity model (equations \ref{eq: true kdv equation}, \ref{eq: true kdv ic and bc} \& \ref{eq: kdv equation analytical solution}). 
The solutions of the two models have the same initial condition, however, their evolutions are drastically different. With the high-fidelity model, the two solitons interact elastically, i.e., their amplitudes and shapes are unchanged after the interaction, however, they do experience a phase shift in their positions. With the low-fidelity model, however, the two solitons do not even come close to interacting with each other. 

\begin{figure}[h]
  \centering
  \includegraphics[width=1\textwidth]{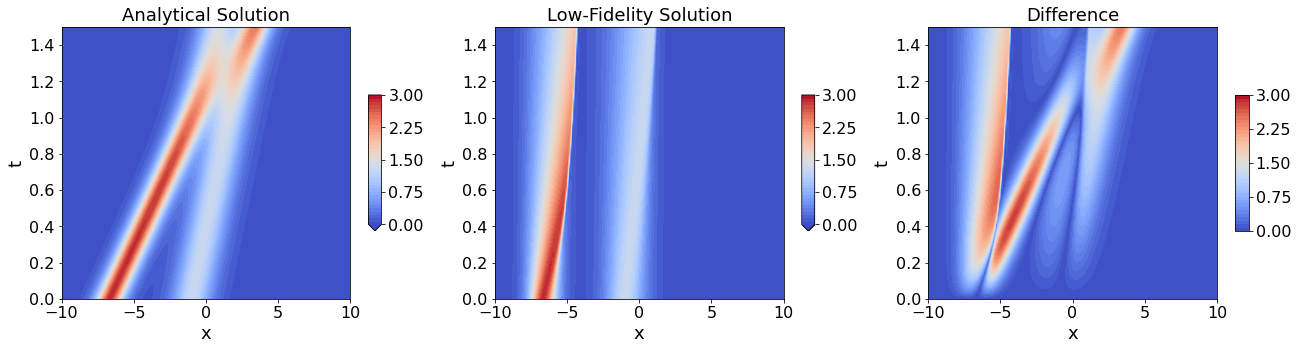}
  \caption{Comparison of the numerical solution of the KdV-Burgers equation with only the advection term (equation \ref{eq: just advection term}; low-fidelity model; \textit{middle plot}), with the analytical solution corresponding to the equation with stronger advection and $3^{rd}$ order derivative term (equations \ref{eq: true kdv equation}, \ref{eq: true kdv ic and bc} \& \ref{eq: kdv equation analytical solution}; high-fidelity model; \textit{left plot}). The low-fidelity model is solved on a grid with $N_x=200$ grid points. The absolute difference between the two solutions is provided in the \textit{right panel}.
  }
  \label{fig: Exp1a_analy_low-fid_diff}
\end{figure}

\textcolor{black}
{\emph{Training: NN architecture, data, and loss function.}}
For the \textit{g}nCM, we only consider the Markovian term with a simple neural network with no hidden layer and only linear activation in the output layer, in-effect equivalent to a linear combination of the inputs. The training data consists of the analytical solution (equation \ref{eq: kdv equation analytical solution}) sampled at time intervals of 0.01 until time $t=1.0$, with a validation period from $1.0\leq t \leq 1.25$. In all the experiments, we use both $\mathcal{L}_1$ and $\mathcal{L}_2$ regularization for the weights of the neural network, and prune them if their value drops below a certain threshold (only if the weightage of $\mathcal{L}_1$ regularization is non-zero), in order to promote sparsity. 
The set of tuned hyperparameters used to generate the results presented next are provided in the supplementary information, section \ref{sec: Hyperparameters}. 
Given the analytical solution data, $\{u^{true}(x, T_i), \; -L \leq x \leq L \}_{i=1}^M$, the loss function is based on time and space averaged mean-absolute-error (MAE), $\mathcal{L} = \frac{1}{M} \sum_{i=1}^M \int_{-L}^L \frac{1}{2L} |u^{pred}(x, T_i) - u^{true}(x, T_i)| dx$, where $M=100$ is the number of high-fidelity solution states at different times available for training. 
%\PFJL{We should give the value of M in the "The training data" sentence, or here.}
%
%\AG{$M$ is a derived hyperparamter, it is not input by the user explicitly. The hyperparamters which determine its value are mentioned in the SI. If really needed, we could mention the $M$ in SI.}
%\PFJL{What I mean is that it is not clear how much data is given. There is a maximum based on the time-step and number of data points at each step. How many time-steps?}
%\AG{Doesn't the line above: ``The training data consists of the analytical solution (equation \ref{eq: kdv equation analytical solution}) sampled at time intervals of 0.01 until time $t=1.0$" tells the same?}

\textcolor{black}{\emph{Learning results.}}
We perform 6 repeats of the experiment with exactly the same set of hyperparameters, and the learned model with the mean and standard deviation of the weights is as follows,
\begin{equation}
\label{eq: learned kdv equation}
    \begin{split}
        \frac{\partial u}{\partial t} = -u\frac{\partial u}{\partial x} - (4.9680 \pm 0.0008) u\frac{\partial u}{\partial x} - (1.0105 \pm 0.0002)\frac{\partial^3 u}{\partial x^3}\,.
    \end{split}
\end{equation}
% kdv_burgers_testcase_nODE/model_dir_case11fc*
% {-1.0106798, -1.0103933, -1.010274, -1.0103426, -1.010876, -1.0105913} -> Mean: -1.0105261666666665; Std: 0.0002102249641588044
% {-4.968391, -4.968655, -4.9688053, -4.966333, -4.9674096, -4.968145} -> Mean: -4.9679564833333325; Std: 0.0008531840411397739
% RMSE: {0.00845003, 0.00452224, 0.00606321, 0.00738099, 0.00669685, 0.00471607} -> Mean: 0.006304898333333334; Std: 0.0013948625205901914
%RMSE Truth: 0.02509284
The true coefficients corresponding to the learned $u\frac{\partial u}{\partial x}$ and $\frac{\p^3 u}{\p x^3}$ terms are $-5.0$ and $-1.0$, respectively. The learned closure is able to recover the true model, and the slight discrepancy in the learned coefficients is to compensate for the very small discretization and truncation errors. To illustrate this, we compare the root-mean-square-error (RMSE), $\mathcal{L} = \frac{1}{M} \sum_{i=1}^M \sqrt{ \sum_{j=1}^{N_x} \frac{1}{N_x} (u^{pred}(x_j, T_i) \allowbreak - u^{true}(x_j, T_i))^2}$, of the learned closure and the true model solved using the same numerical schemes. The RMSE (mean and standard deviation) obtained for the learned closure and the true model solved numerically is $0.0063 \pm 0.0014$ and $0.0251$, respectively. Thus, on average, the learned closure leads to a smaller RMSE than the error of the numerically-solved true model. 
We note that this excellent accuracy in the coefficients of the recovered (learned) model compared to the true model is similar to that observed in SINDy and its variants for the KdV PDE in \cite{rudy2019data, messenger2021weak, both2021deepmod}.

\textcolor{black}{\emph{Sensitivity.}}
The learning was sensitive to batch-time, and higher values were especially detrimental to convergence. This behavior is in general observed when the error between the low- and high-fidelity models is large, e.g., when there is no low-fidelity model. 
%
%\PFJL{We could cite Aman's paper in prep here where more of this is studied.} \AG{Done.}
%
Using a smaller batch size and regularization weights lead to slightly different values of the learned coefficients. 
This is especially noted for the $u^2\frac{\p u}{\p x}$ term, whose weight tends towards a non-zero value with a very small magnitude. 
For a study on the impact of different hyperparameters (encountered specifically in the nCM framework and SciML in general) on training, we refer to \cite{jalan_etal_2023}.
%\AG{The reader is further pointed toward \cite{jalan_etal_2023}, which provides a detailed and comprehensive study on the impact of different hyperparameters (encountered specifically in the nCM framework and SciML in general) on training.}
%
In the current set of experiments, the learning framework is able to recover the known true model and, due to this, we do not additionally focus on demonstrating generalization over initial conditions, boundary conditions, and grid resolution.

\subsection{Experiments 1b: Advecting Shock - Model Discovery and Generalization}
%auto-ignore
\label{sec: Experiments 1b}
%\textcolor{blue}{Summ.:}
In the second set of experiments, we employ the advecting shock PDE models. First, a \textit{g}nCM discovers the leading truncation term and a correction to the nonlinear advection term by interpreting the learned Markovian neural closure. Second, we utilize both Markovian and non-Markovian \textit{g}nCM terms trained on data corresponding to just a few combinations of grid resolution and Reynolds number, and demonstrate the generalization of the learned closure model over grid resolution, Reynolds number, initial and boundary conditions, along with superior performance compared to the popular Smagorinsky closure model. We further interpret the learned closure by analysing the weights of the learned neural networks, and find the closure to be independent of the Reynolds number despite it being one of the functional inputs.

\textcolor{black}
{\emph{Setup: True model, data generation, and low-fidelity model.}}
We consider the classic form of the Burgers equation as the governing  high-fidelity model,
\begin{equation}
\label{eq: classic Burger's eq}
    \begin{split}
        \frac{\p u}{\p t} = -u\frac{\p u}{\p x} + \nu\frac{\p^2 u}{\p x^2} \;, \quad 0 \leq x \leq L, ~ t \in (0, T] \;,
    \end{split}
\end{equation}
where $\nu$ is the diffusion coefficient. 
The data is generated from an analytical solution of this Burgers equation  \ref{eq: classic Burger's eq} with initial and boundary conditions,
\begin{equation}
\label{eq: classic Burger's eq ICs and BCs}
\begin{split}
   u(x, 0) &= \frac{x}{1 + \sqrt{\frac{1}{t_0}} \exp\left(Re\frac{x^2}{4} \right)}\,, \quad  u(0, t) = 0, \quad \text{and} \quad \frac{\partial u(x, t)}{\partial x}\bigg|_{x=L} = 0 \,,
   \end{split}
\end{equation}
where the Reynolds number $Re = 1/\nu$ and $t_0 = \exp(Re/8)$. This solution is given by,
\begin{equation}
\label{eq: Analytical solution for Burger's}
    \begin{split}
         u(x, t) &= \frac{x / (t+1)}{1 + \sqrt{\frac{t+1}{t_0}} \exp\left(Re\frac{x^2}{4t + 4} \right)} \,.
    \end{split}
\end{equation}
However, when the discrete version of the above equation \ref{eq: classic Burger's eq} is solved numerically, truncation and round-off errors occur and the numerical solution incurs discretization errors \cite{chapra2011numerical,lermusiaux_2.29_notes}.
%
%\PFJL{Here, I wonder if we need all these definitions. Also, the discretization error (error in solution) is different than the truncation error (error of the discrete PDE). We may need to do corrections everywhere.}
%\AG{I don't think we need these definitions, however, we should make a passing remark to truncation and discretization errors. I changed discretization to truncation error everywhere.}
%
% However, when the discrete version of the above equation \ref{eq: classic Burger's eq} is solved numerically, the numerical solution incurs errors from three sources;
% \begin{enumerate*}
%     \item \textit{Projection error}, which accounts for the fact that the exact solution is approximated using a finite number of degrees of freedom. This error cannot be avoided;
%     \item \textit{Discretization error}, which accounts for the fact that partial derivatives which appear in the continuous problem are approximated on the computational grid using Finite Difference, Finite Volume, Finite Element (or other similar) schemes;
%     \item \textit{Resolution error}, which accounts for the fact that the absence of some scales of the exact solution results in the evaluation of the non-linear flux function to be inexact, even if the discretization error is driven to zero \cite{sagaut2006large}. 
% \end{enumerate*}
%
%\PFJL{How can we have a nonzero resolution error if the truncation error is zero? I guess if we use the analytical solution to evaluate fluxes on a discrete grid? I need to check Sagaut and his examples.}
%

\textcolor{black}{\emph{Numerics.}}
We solve the Burgers equation (\ref{eq: classic Burger's eq}) numerically with the following schemes:
%we use a finite difference scheme. Specifically, 
$1{st}$ order accurate upwind for the advection term, $2^{nd}$ order accurate central-difference for the diffusion term, and \textit{Vode} scheme for adaptive time-stepping. 
Thus, the leading order truncation error term is given by, $-\frac{\Delta x}{2} u \frac{\partial^2 u}{\partial x^2} + \mathcal{O}(\Delta x^2)$, where $\Delta x$ is the uniform grid-spacing. The terms in $\mathcal{O}(\Delta x^2)$ contain spatial derivatives of order 3 and above.
\textcolor{black}{\emph{Comparison LF-HF:}}
A comparison of the analytical (equation \ref{eq: Analytical solution for Burger's}) and numerical solution of the Burgers equation is provided in Figure \ref{fig: Exp1b_Analy_vs_LowRes_Nx50_Re1000_NBC}. One can clearly notice the effects of numerical diffusion and the error in the location of the shock peak at later times due to
%discretization and 
truncation
errors.
%
%\PFJL{We likely need to describe Fig 3 here, as we do in 3.2.1? We don't describe it anywhere.} \AG{Done.}

\begin{figure}[h]
  \centering
  \includegraphics[width=1\textwidth]{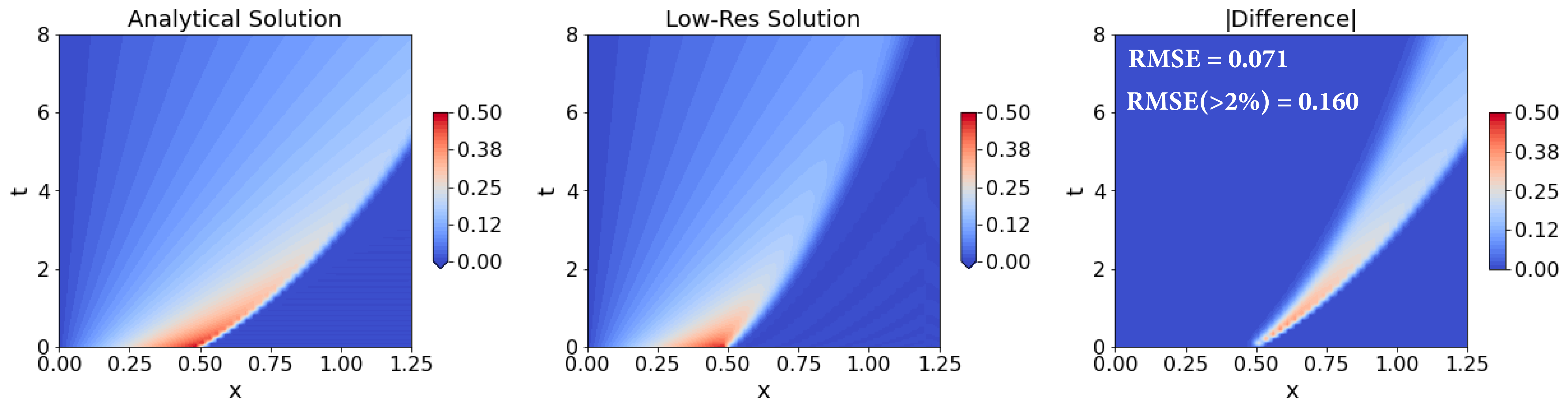}
  \caption{Comparison of the numerical solution of the Burgers equation (with $Re=1000$) on a low-resolution grid  (equations \ref{eq: classic Burger's eq} \& \ref{eq: classic Burger's eq ICs and BCs}; low-fidelity model; \textit{middle plot}), with its corresponding analytical solution (equation \ref{eq: Analytical solution for Burger's}; high-fidelity model; \textit{left plot}). The low-fidelity model is solved on a grid with $N_x=50$ grid points, and the absolute difference between the two solutions is provided in the \textit{right plot}. We also provide a pair of time-averaged errors, specifically: root-mean-squared-error (RMSE); and RMSE considering only the grid points where the error is at least $2\%$ of the maximum velocity value, denoted by RMSE($>2\%$).
  }
  \label{fig: Exp1b_Analy_vs_LowRes_Nx50_Re1000_NBC}
\end{figure}

\subsubsection{Learning interpretable truncation errors and nonlinear flux corrections}
\label{sec:learn_trunc}

\textcolor{black}
{\emph{Training: NN architecture, data, and loss function.}}
First, we only consider a Markovian closure term based on a library composed of second-degree combinations of $u$, $\frac{\partial u}{\partial x}$, and $\frac{\partial^2 u}{\partial x^2}$.
The library explicitly omits $u\frac{\partial u}{\partial x}$ because it is already known as part of the governing equation, and $u^2$ because it cannot be part of truncation error due to the absence of any derivative.  
Hence, the Markovian closure term is assumed to be a linear combination of $\left\{\right. \Delta x \left( \frac{\partial u}{\partial x} \right)^2, \allowbreak \Delta x^3 \left( \frac{\partial^2 u}{\partial x^2} \right)^2, \allowbreak \Delta x^2 \left(\frac{\partial u}{\partial x} \frac{\partial^2 u}{\partial x^2} \right), \allowbreak \Delta x \left( u \frac{\partial^2 u}{\partial x^2} \right) \left.\right\}$, of which the fourth term is true but unknown leading order truncation error term itself, 
and other terms are informed but still expected to be redundant.
% First, we only consider a Markovian closure, as a linear combination of a library of four terms, $\left\{\right. \Delta x \left( \frac{\partial u}{\partial x} \right)^2, \allowbreak \Delta x^3 \left( \frac{\partial^2 u}{\partial x^2} \right)^2, \allowbreak \Delta x^2 \left(\frac{\partial u}{\partial x} \frac{\partial^2 u}{\partial x^2} \right), \allowbreak \Delta x \left( u \frac{\partial^2 u}{\partial x^2} \right) \left.\right\}$, out of which three are up to second-degree combinations of $\frac{\partial u}{\partial x}$ and $\frac{\partial^2 u}{\partial x^2}$, and the fourth is the leading order truncation error term itself. 
%
%\PFJL{Can we justify the library better, as we did in 3.1? Is it a complete one?} \AG{Done.}
%
Each of the terms is multiplied with appropriate powers of $\Delta x$, such that the closure terms are dimensionally consistent with the other terms of the Burgers equation.
The $4^{th}$ order accurate central and upwind finite-difference schemes \cite{rahul2006one} are used to compute the spatial derivatives in the Markovian closure, so as to eliminate additional sources of truncation error from our analysis.
The training data consists of the analytical solution up until $T=4.0$ solved in a domain of length $L=1.25$ and saved at every 0.01 time-intervals, for three different combinations of $N_x$ (number of grid points in $x-$direction) and $Re$. The chosen ($N_x$, $Re$) pairs, $\{(100, 50), (150, 750), ~\text{and} ~ (200, 1250) \}$, are such that the $-\frac{\Delta x}{2} u \frac{\partial^2 u}{\partial x^2}$ term is really the leading source of error. In every epoch, we parse through the training data of each of these pairs, selected in random order by sampling without replacement. We tune the hyperparameters based on performance in the training period ($0.0 \leq t \leq 4.0$) and the validation period ($4.0 \leq t \leq 6.0$), and these are provided in Section \ref{sec: Hyperparameters}. 
The Markovian closure model is a simple neural network with no hidden layers and only linear activation in the output layer, in-effect equivalent to a linear combination of the inputs. Given the analytical solution, $\{u^{true}(x, T_i), \; 0 \leq x \leq L \}_{i=1}^M$, the loss function is once again the time and space averaged mean-absolute-error (MAE), $\mathcal{L}= \frac{1}{M} \sum_{i=1}^M \int_{0}^L \frac{1}{L} |u^{pred}(x, T_i) - u^{true}(x, T_i)| dx$, where $M=400$ is the number of high-fidelity solution states at different times available for training. 
%
%\PFJL{We should give the value of M in the "The training data" sentence, or here.}

\textcolor{black}{\emph{Learning results.}}
We perform 8 repeats of the same experiment with the tuned hyperparameters. The resulting learned model with the mean and standard deviation of the coefficients is as follows,
% burgers_testcase_nODE_v2/model_dir_case1*
% 1st term: {0.14284897, 0.12194626, 0.12643105, 0.14212216, 0.14560033, 0.1540051, 0.09875096, 0.13081735} -> mean: 0.1328; std: 0.01742
% 2nd term: {0.0, 0.06395595, -0.01544803, 0.01705859, 0.0, 0.0, 0.0, 0.0100266 } -> mean: 0.009449; std: 0.02392
% 4th term: {-0.31270245, -0.3098588, -0.3444906, -0.30822447, -0.3108834, -0.30413023, -0.3674688, -0.32524246} -> mean: -0.3229; std: 0.02218
% -1st term + 4th term: {-0.14284897-0.31270245, -0.12194626-0.3098588, -0.12643105-0.3444906, -0.14212216-0.30822447, -0.14560033-0.3108834, -0.1540051-0.30413023, -0.09875096-0.3674688, -0.13081735-0.32524246} -> mean: -0.4557; std: 0.01164
\begin{equation}
    \begin{split}
         & \mathcal{F}_{NN} \left(  \Delta x \left( \frac{\partial u}{\partial x} \right)^2, \Delta x^3 \left( \frac{\partial^2 u}{\partial x^2} \right)^2, \Delta x^2 \left(\frac{\partial u}{\partial x} \frac{\partial^2 u}{\partial x^2} \right), \Delta x \left( u \frac{\partial^2 u}{\partial x^2} \right); \phi \right) \\
          & \hspace{0.15\textwidth}=  (0.133 \pm 0.017) \Delta x \left( \frac{\partial u}{\partial x} \right)^2 + (0.009 \pm 0.023)\Delta x^3 \left( \frac{\partial^2 u}{\partial x^2} \right)^2 \\
        & \hspace{0.25\textwidth} - (0.323 \pm 0.022) \Delta x \left( u \frac{\partial^2 u}{\partial x^2} \right) \,.
    \end{split}
    \label{eq: Leaned Markovian closure Burger's equation}
\end{equation}
% MB: Why are there two equations above?
%
For evaluation, we first compare the performance of this learned \textit{g}nCM w.r.t.\ using the true leading truncation error term ($-\frac{\Delta x}{2} u \frac{\partial^2 u}{\partial x^2}$) as the closure itself. 
For both cases, we evolve the Burgers equation with the respective closure terms up until $T=8.0$ (beyond training and validation time-periods), for 35
%\PFJL{add how many here} \AG{Done} 
$(N_x, Re)$ pairs in the 2D domain spanned by $50 \leq N_x \leq 200$ and $50 \leq Re \leq 1500$.
In Figure \ref{fig: NxvsRe plots} we provide the RMSE$(>2\%)$ error (see Figure \ref{fig: Exp1b_Analy_vs_LowRes_Nx50_Re1000_NBC} for description). 
When the true leading truncation error term is used as the closure, we find that increasing $Re$ and lowering $N_x$ values leads to instabilities in the solution which causes it to explode. 
On the contrary, in the learned \textit{g}nCM case, even though it was not shown any training data in the high $Re$ and low $N_x$ regime, it still provides a stable solution, and, on average, performs better than its counterpart in the other regions of the $(N_x, Re)$ domain. 
To interpret the learned closure further, we rewrite it by substituting, $\frac{\partial }{\partial x}\left( u\frac{\partial u}{\partial x} \right) = \left( \frac{\partial u}{\partial x} \right)^2 + \left( u \frac{\partial^2 u}{\partial x^2} \right)$ in equation \ref{eq: Leaned Markovian closure Burger's equation},
% burgers_testcase_nODE_v2/model_dir_case1*
\begin{equation}
    \begin{split}
         & \mathcal{F}_{NN} \left(  \Delta x \left( \frac{\partial u}{\partial x} \right)^2, \Delta x^3 \left( \frac{\partial^2 u}{\partial x^2} \right)^2, \Delta x^2 \left(\frac{\partial u}{\partial x} \frac{\partial^2 u}{\partial x^2} \right), \Delta x \left( u \frac{\partial^2 u}{\partial x^2} \right); \phi \right) \\
          & \hspace{0.15\textwidth} =  (0.133 \pm 0.017) \Delta x \frac{\partial }{\partial x}\left( u\frac{\partial u}{\partial x} \right) + (0.009 \pm 0.023)\Delta x^3 \left( \frac{\partial^2 u}{\partial x^2} \right)^2 \\
         & \hspace{0.25\textwidth} - (0.456 \pm 0.012) \Delta x \left( u \frac{\partial^2 u}{\partial x^2} \right) \,.
    \end{split}
\end{equation}
Thus, the learned \textit{g}nCM contains the $\Delta x \left( u \frac{\partial^2 u}{\partial x^2} \right)$ term with a coefficient of correct sign but slightly smaller value -- in absolute value -- in comparison to that of the true leading truncation error term. Along with that, the other significant term, $\Delta x \frac{\partial }{\partial x}\left( u\frac{\partial u}{\partial x} \right)$, corresponds to a first-order Taylor series correction to the nonlinear advection term, and can help with mitigating the resolution error highlighted earlier.
Finally, it is remarkable that the important $\Delta x \frac{\partial }{\partial x}\left( u\frac{\partial u}{\partial x} \right)$ term was missing from the input features; however, to our surprise, it is still accounted for indirectly in the learned closure, utilizing the redundant terms present in the input feature library. This highlights a noteworthy learning capability of the \textit{g}nCM.
%

% burgers_testcase_nDistDDE_v2/model_dir_case5f and 5fa*
\begin{figure}[h]
	\centering
	\subfloat[][Leading truncation error term as closure]{\includegraphics[width=0.45\textwidth]{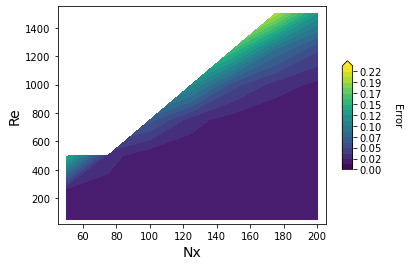}} \qquad
	\subfloat[][\textit{g}nCM with only Markovian closure term]{\includegraphics[width=0.45\textwidth]{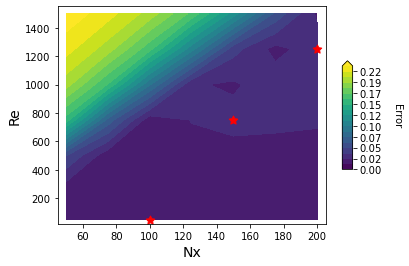}} \\
	\subfloat[][Smagorinsky closure]{\includegraphics[width=0.45\textwidth]{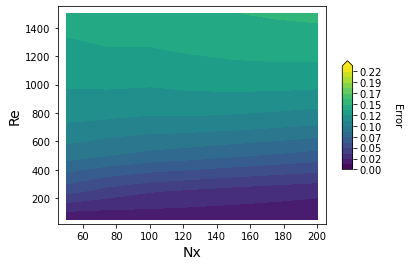}} \qquad
	\subfloat[][\textit{g}nCM\label{fig: NxvsRe plots for full gnCM closure}]{\includegraphics[width=0.45\textwidth]{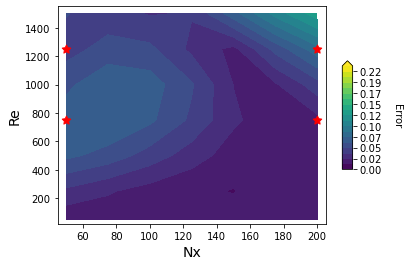}}
	\caption{Performance of four closure models for the Burgers equation (equations \ref{eq: classic Burger's eq} \& \ref{eq: classic Burger's eq ICs and BCs})  evaluated for various $(N_x, Re)$ pairs in the 2D domain spanned by $50 \leq N_x \leq 200$ and $50 \leq Re \leq 1500$. 
    The error provided is the $RMSE\,(>2\%)$ (see Figure \ref{fig: Exp1b_Analy_vs_LowRes_Nx50_Re1000_NBC} for description) computed w.r.t.\ the corresponding analytical solutions (equation \ref{eq: Analytical solution for Burger's}) for $0.0\leq t \leq 8.0$ in a domain of length $L = 1.25$. 
    \textit{(a):} Leading truncation error term, $-\frac{\Delta x}{2} u \frac{\partial^2 u}{\partial x^2}$, as closure. The \textit{white} region in the top-left denotes an unconverged numerical solution; 
    \textit{(b):} Learned  \textit{g}nCM with only the Markovian term, with the three \textit{red} $\star$'s marking the ($N_x, Re$) pairs used as training data.; 
    \textit{(c):} Smagorinsky LES model with $C_s = 1.0$; \textit{(d):} Learned \textit{g}nCM with both Markovian and non-Markovian closure terms, with the four \textit{red} $\star$'s marking the ($N_x, Re$) pairs used as training data.}
	\label{fig: NxvsRe plots}
\end{figure}

\subsubsection{Learning generalizable and interpretable closures}

\textcolor{black}
{\emph{Training: NN architecture, data, and loss function.}}
Keeping the Markovian closure term formulation of Section \ref{sec: Hyperparameters}, we now add the non-Markovian closure term with inputs, $\{u, \frac{\partial u}{\partial x}, \frac{\partial^2 u}{\partial x^2}, \nu, \Delta x\}$, discretized using $4^{th}$ order finite-difference schemes, and the deep-NN architecture given in Table \ref{table: All architecture}. 
We utilize a fully-connected deep-NN with four hidden-layers and the non-linear \textit{swish} activation.  
The output of the NN is multiplied with $|u|$ to ensure that the contribution of the non-Markovian closure term is zero in the right-hand parts of the domain where the shock is yet to reach. 
As the non-Markovian closure term is nonlinear, we do not explicitly make the inputs dimensionally consistent with other terms in the Burgers equation. The overall training and evaluation setup are as in Section  \ref{sec:learn_trunc}, however, this time four pairs of $(N_x, Re)$ are used such that all four combinations of high and low $N_x$ and $Re$ are contained in the training data. The chosen pairs were, $\{(50, 750), (200, 750), (50, 1250), (200, 1250)\}$. 
The tuned set of hyperparameters is provided in Section \ref{sec: Hyperparameters}. The time-delay, $\tau = 0.075$, is based on the optimal-time delay established for the Burgers equation experiments in \cite{gupta_lermusiaux_PRSA2021}. 

\textcolor{black}
{\emph{Learning results.}}
We perform 7 repeats of the experiment with exactly the same set of tuned hyperparameters. The learned coefficients for the Markovian term are different than those in equation \ref{eq: Leaned Markovian closure Burger's equation} due to the presence of the non-Markovian term, however, once again, the most weightage is given to the $\Delta x \left( \frac{\partial u}{\partial x} \right)^2$ and $\Delta x \left( u \frac{\partial^2 u}{\partial x^2} \right)$ terms. 
Upon inspection, the weights of the input layer of the deep-NN in the non-Markovian term being multiplied with $\nu$ were consistently found to be particularly small ($\mathcal{O}(10^{-4})$), indicating that the learned closure is independent of $\nu$.
For one of the experiment runs, 
we show in Figure \ref{fig: NxvsRe plots} the performance for $(N_x, Re)$ pairs in the 2D domain spanned by $50 \leq N_x \leq 200$ and $50 \leq Re \leq 1500$, and compare it with that of the popular Smagorinsky model used for subgrid-scale turbulence closure in large eddy simulations (LES). To the Burgers equation (\ref{eq: classic Burger's eq}), this model introduces a dynamic turbulent eddy viscosity ($\nu_e$) resulting in,
\begin{equation}
\label{eq: classic Burger's eq with smagorinsky}
    \begin{split}
        \frac{\p u}{\p t} = -u\frac{\p u}{\p x} + \nu\frac{\p^2 u}{\p x^2} + \frac{\partial}{\partial x}\left(\nu_e\frac{\partial u}{\partial x}\right)\;,
    \end{split}
\end{equation}
where $\nu_e = (C_s \Delta x)^2 \big| \frac{\partial u}{\partial x}\big|$ and $C_s$ is the Smagorinsky constant. 
As the rectangle formed by the training $(N_x, Re)$ pairs is only a subset of the rectangle in which we evaluate the learned closure, we are testing both interpolation and extrapolation performance w.r.t.\ changing the physical parameter governing the model and grid resolution. 
We find that the learned \textit{g}nCM clearly outperforms the Smagorinsky model. 
It should be noted, that in Figure~\ref{fig: NxvsRe plots for full gnCM closure}, the bottom-right corner (low $Re$ and high $N_x$ region) has inherently small errors even without the presence of a closure. Further, the amount of error between low-fidelity and high-fidelity solutions is different for the four training data $(N_x, Re)$ combinations; for example, $(50, 1250)$ (coarsest resolution, higher Re) should incur the most error. 
Thus, we notice a differential in the impact of learned \textit{g}nCM in reducing the error at and around different training data $(N_x, Re)$ pairs.

As claimed earlier, we expect the learned \textit{g}nCM to be also generalizable over different boundary conditions. We tested this by modifying the boundary conditions. The analytical solution (equation \ref{eq: Analytical solution for Burger's}) used in training corresponded to Neumann boundary conditions on the right edge of the domain. This was changed to a zero Dirichlet boundary condition. Furthermore, the length of the domain was decreased to $L=1$, and $N_x = 50$ number of equally-spaced grid points were used in our low-fidelity model with $Re=1000$. Since no closed-form analytical solution exists for the Dirichlet boundary conditions case, we solve the system with $N_x = 1000$ grid points and use that as the true solution for comparing the performance of our learned closure.  
In Figure \ref{fig: LowRes_Nx50_Re1000_DBC}, we find that the learned \textit{g}nCM is able to keep the errors remarkably low throughout the time period encompassing training, testing, and prediction.

\begin{figure}[h]
	\centering
	\subfloat[][No closure]{\includegraphics[width=0.75\textwidth]{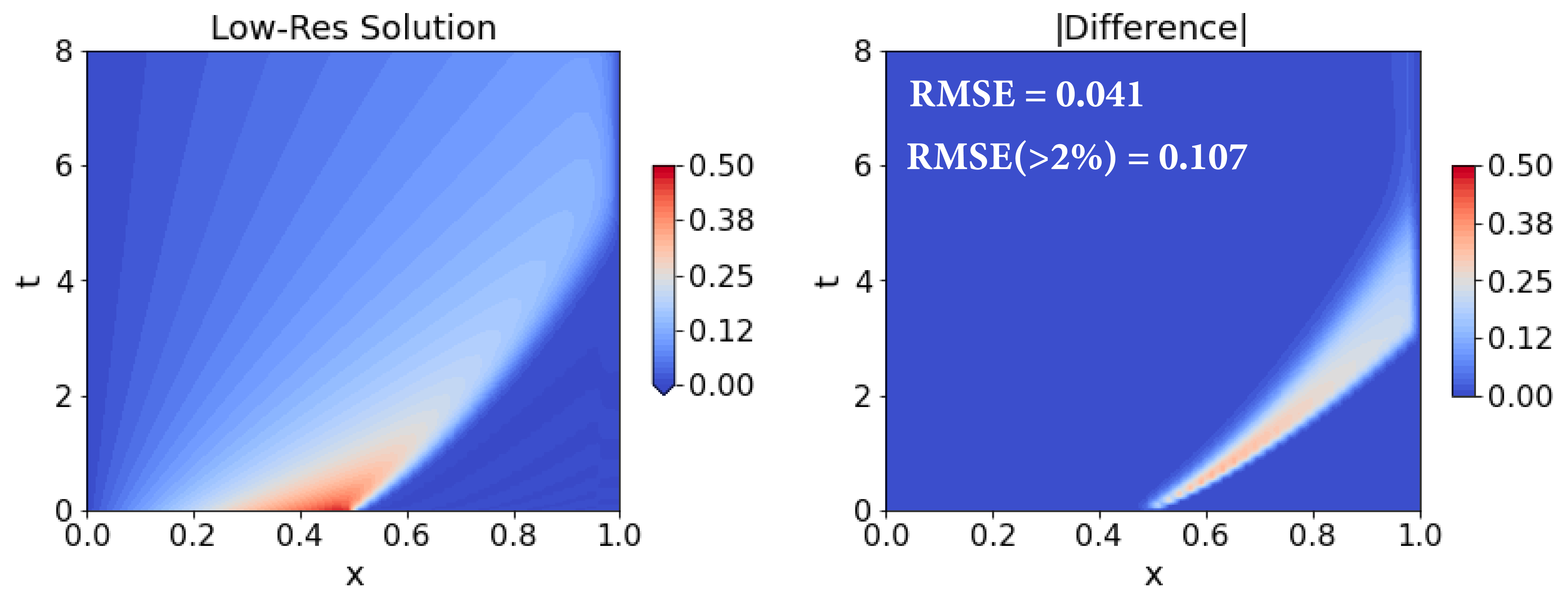}} \\
	\subfloat[][\textit{g}nCM]{\includegraphics[width=0.75\textwidth]{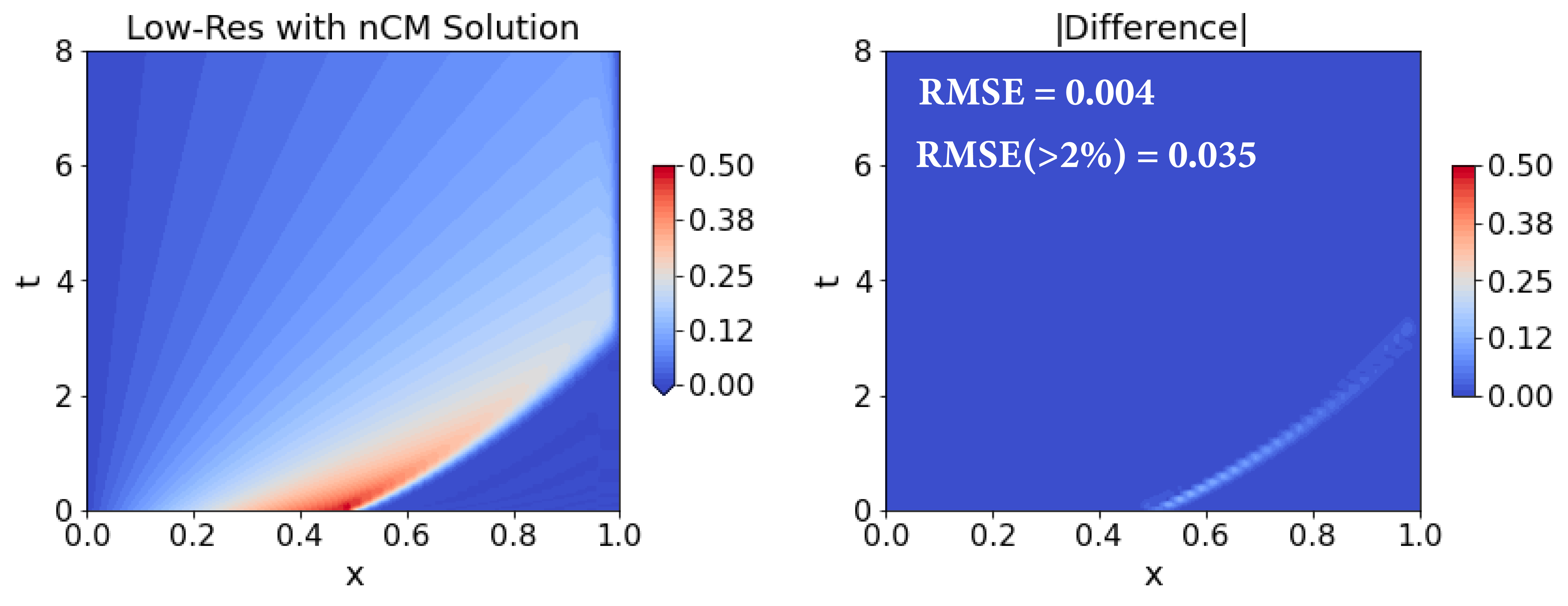}}
	\caption{Solution of the Burgers equation with and without the learned generalized neural closure model (\textit{g}nCM) for $Re = 1000$, a low-resolution grid ($N_x = 50$), and zero Dirichlet boundary condition on the right edge. For each case, we also provide the pair of time-averaged errors (see Figure \ref{fig: Exp1b_Analy_vs_LowRes_Nx50_Re1000_NBC} for description).}
	\label{fig: LowRes_Nx50_Re1000_DBC}
\end{figure}

\textcolor{black}
{\emph{Sensitivity.}}
In general, the quality of learning was less sensitive to the batch-time hyperparameter, however, higher values led to more interpretable closures. Using lower-order finite-difference schemes for the closure inputs did not compromise the performance of the learned closures, however, it did lead to a decrease in interpretability. Sensitivity to other hyperparameters was similar to that observed in Experiments-1a.

\subsection{Experiments 2a: Ocean Acidification - Interpretable Model Discrimination}
\label{sec: Experiments 2a}
%auto-ignore
%\textcolor{blue}{Summ.:}
In the third set of experiments, we consider coupled physical-biological-carbonate PDE models that are used to study ocean acidification (OA). We utilize the Markovian neural closure model to interpretably discriminate between candidate functional forms of uncertain Zooplankton mortality term.

\textcolor{black}
{\emph{Setup: True model, data generation, and low-fidelity model.}}
% Next, we will use our framework to determine the functional form of certain processes in ocean acidification (OA) models. 
OA models are used to study and predict essential carbonate chemistry and biological production cycles, and their interplay with global warming.
A plethora of biogeochemical models have been proposed. They differ in their complexity, or ability to resolve different biological processes. A set of parameter values and functional forms that might work in a particular ocean region may not apply anywhere else. As an additional source of complexity, there may be seasonal variability in these functional forms \cite{lermusiaux_et_al_CS2004,tian_etal_2004,gupta_lermusiaux_PO2022}.

For this set of experiments, the high-fidelity model is similar to the Hadley Centre Ocean Carbon Cycle (HadOCC) model \cite{palmer2001production}, where the biological part is a modified version of four components (nutrients (N), phytoplankton (P), zooplankton (Z), and detritus (D)) developed in 
%by Tian \textit{et al.}
\cite{tian2015model} for the Gulf of Maine, along with dissolved inorganic carbon (DIC) and total alkalinity (TA) for the carbonate part. The NPZD model is,
\begin{equation}
\label{eq: gnCM NPZD model}
        \begin{split}
            \frac{dN}{dt} &= -U_P + \lambda G_Z + \varepsilon  D \;, \\
            \frac{dP}{dt} &= U_P - G_Z - m_P P \;, \\
            \frac{dZ}{dt} &= \gamma G_z - M_Z(Z) \;, \\
            \frac{dD}{dt} &= (1-\gamma-\lambda) G_Z + m_P P + M_Z(Z) - \varepsilon D \;,
        \end{split}
\end{equation}
where $U_P$ is the phytoplankton growth, regulated by nitrogen limitation based on Michaelis-Menten kinetics ($f(N)$) and photosynthetically active radiation ($f(I)$), and $G_Z$ the zooplankton grazing, each given by,
\begin{equation}
    \begin{split}
        U_P &= \mu_{max} f(N) f(I) P, \quad f(N) = \frac{N}{N+K_N}, \\
        f(I) &= (1-\exp(\alpha I / \mu_{max})) \exp(-\beta I / \mu_{max}) \\
            I(z) &= I_0 \exp(-k_W z), \quad G_Z = \frac{g_{max} ZP^2}{P^2 + K_P^2}\;,
    \end{split}
\end{equation}
and $M_Z(Z)$ is the to-be-learned zooplankton mortality. 
In these equations, the concentration of biological variables is measured in nitrogen ($mmol~N~m^{-3}$), 
$z$ is depth, and the other parameters are: $\mu_{max}$, maximum growth rate of phytoplankton; 
$K_N$, half-saturation constant; $\alpha$ and $\beta$, light-growth slope and inhibition coefficient; 
$I_0$, photosynthetically active radiation (PAR) at the sea surface; 
$k_W$, attenuation coefficient of water;
$g_{max}$, zooplankton maximum grazing rate; $K_P$, half-saturation constant for zooplankton grazing; 
$\gamma$, assimilation coefficient; $m_z$, zooplankton mortality coefficient; 
$m_p$, phytoplankton mortality coefficient; $\lambda$, active respiration zooplankton expressed as a fraction of grazing; and $\varepsilon$, remineralization rate of detritus. The carbon in the system is coupled with the nitrogen by fixed carbon-nitrogen ratios, $C_P$, $C_Z$, and $C_D$,
\begin{equation}
\label{eq: gnCM OA ODE model}
        \begin{split}
            \frac{d (DIC)}{dt} &= -C_P \frac{dP}{dt} - C_Z \frac{dZ}{dt} - C_D \frac{dD}{dt} - \gamma_c C_P U_P \;,  \\
            \frac{d(TA)}{dt} &= -\frac{1}{\rho_w}\frac{dN}{dt} - \frac{2\gamma_c C_P U_P}{\rho_w} \,, 
        \end{split}
\end{equation}
and neither DIC nor TA has any effect on the biology because phytoplankton growth is not carbon limited. The last term in the DIC equation represents the precipitation of calcium carbonate to form shells and other hard body parts, which subsequently sink below the euphotic zone, also known as ``hard flux''. This flux is modeled to be proportional (and additional) to the uptake of carbon for primary production.
The chemistry dictates the decrease in total alkalinity by two molar equivalents for each mole of carbonate precipitated. In general, since TA is measured in $mmol ~C~ kg^{-1}$ (or $\mu mol ~C~ kg^{-1}$), we divide the right-hand-side (RHS) of the TA equation by the density of sea-water ($\rho_w$). Moreover, the units of DIC concentration are $mmol ~C~m^{-3}$.

The above biological and carbonate models are often coupled with physical models to introduce both spatial and temporal components. For our experiments, we use a 1-D diffusion-reaction PDE with vertical eddy mixing parameterized by the operator $\partial/\partial z \left (K_z(z, M)\partial/\partial z (\bullet) \right)$, where $K_z$ is a dynamic eddy diffusion coefficient.  A mixed layer of varying depth ($M = M(t)$) is used as a physical input to the OA models. Thus, each biological and carbonate state variable $B(z, t)$ is governed by the following non-autonomous PDE,
\begin{equation}
\label{eq: gnCM ADR eqn}
    \frac{\partial B}{\partial t} = S^B + \frac{\partial}{\partial z}\left( K_z(z, M(t))\frac{\partial B}{\partial z}\right) \;,
\end{equation}
\begin{equation}
    K_z(z, M(t)) = K_{z_b} + \frac{(K_{z_0} - K_{z_b})(\arctan(-\gamma_t (M(t) - z)) - \arctan(-\gamma_t (M(t) - D_z)))}{\arctan(-\gamma_t M(t)) - \arctan(-\gamma_t (M(t) - D_z))} \; ,
\end{equation}
where $K_{z_b}$ and $K_{z_0}$ are the diffusion at the bottom and surface, respectively, $\gamma_t$ is the thermocline sharpness, and $D_z$ is the total depth.
The 1-D model and parameterizations are adapted from 
%Eknes and Evensen, 2002
\cite{eknes2002ensemble} and
%Newberger et. al., 2003 
\cite{newberger2003analysis}. They simulate the seasonal variability in upwelling, sunlight, and  biomass vertical profiles.  The dynamic mixed layer depth, surface photosynthetically-available radiation $I_0(t)$, and biomass fields $B(z, t)$ are shown in Figure~\ref{fig: Exp2a_NPZD-OA_Sq_Mort_Missing}. 
The radiation $I_0(t)$ and total biomass concentration, $T_{bio}(z,t)$, affects $S^B$ and the initial conditions.

%\PFJL{Need to fix such sentences everywhere.} \AG{Done.}
%\PFJL{In this para, up to  Comparison LF-HF, we need to better separate the true high-fidelity model text from the low-fidelity text. This may mean moving some text below in the parts above for the high-fidelity model. We also need to say what is the data (see prior sections where I already did the organization.} 
%\AG{I think what is needed was to move the markovian library in the "Training: NN architecture, data, and loss function" subsection.}
%

To generate data, we first initialize the $N$ state with the depth-varying total biomass concentration and the $P$, $Z$, and $D$ states with zero concentrations, and then run a one-month spin-off of just the NPZD model without the diffusion term and a constant sea-surface solar radiation in order to determine the stable equilibrium of the biological states. These equilibrium states form the initial conditions for the respective states in the NPZD-OA model. To initialize $DIC$, we multiply the equilibrium state for $N$ with the nitrogen-to-carbon ratio that is considered nearly equal to the value of $C_P$. $TA$ is often assumed to have a dependence on salinity and biological processes \cite{artioli2012carbonate}.
The contribution from salinity ($S$ in $PSU$) is modeled using a linear relationship optimized for the Gulf of Maine, $TA = \begin{cases} (198.10 + 61.75 S)/1000 \;, & S < 32.34 \\ (744.41 + 44.86 S)/1000 \;, & S \geq 32.34 \end{cases}$ (Dr.\ P.J.\ Haley Jr., \textit{pers.\ comm.}), while the biological impact is given by equation \ref{eq: gnCM OA ODE model}. We assume a stationary salinity profile described using a sigmoid function $S(z) = A + \frac{K - A}{(C + Q \exp(-Bz))^{1/\nu}}$ with $A = 31.4~PSU$, $K = 32.8~PSU$, $C = 1.0$, $Q = 0.5$, $B = 0.25$, and $\nu = 2.0$. Thus, we can initialize TA based on salinity and evolve it using equation \ref{eq: gnCM OA ODE model} coupled with equation \ref{eq: gnCM ADR eqn}.  

For the low-fidelity model, we assume that we have only prior knowledge about the existence of a linear zooplankton mortality term, i.e., $M_Z(Z) = \frac{m_Z}{2} Z$.
For the high-fidelity model, however,  
the true zooplankton mortality contains an additional quadratic dependence, i.e., $M_Z(Z) = \frac{m_Z}{2} (Z + Z^2)$.

\textcolor{black}{\emph{Numerics.}}
We use a $2^{nd}$ order central difference scheme for the spatial discretization ($N_z = 20$), and \textit{dopri5} \cite{hairer1993solving} scheme for time integration with adaptive time-stepping. \textcolor{black}{\textit{Comparison LF-HF:}} 
In Figure \ref{fig: Exp2a_NPZD-OA_Sq_Mort_Missing}-left- and -mid-columns, we provide a year-long simulation for the NPZD-OA model with quadratic (truth) and linear (prior) $Z$ mortality terms, respectively. We notice the low $Z$ concentration and enhanced $P$ bloom in the former case. 
Figure \ref{fig: Exp2a_NPZD-OA_Sq_Mort_Missing}-right-column provides the absolute difference between the two cases. Values of the model parameters are provided in section \ref{sec: gnCM Experimental Setup}.

\begin{figure}[h!]
  \centering
  \includegraphics[width=0.775\textwidth]{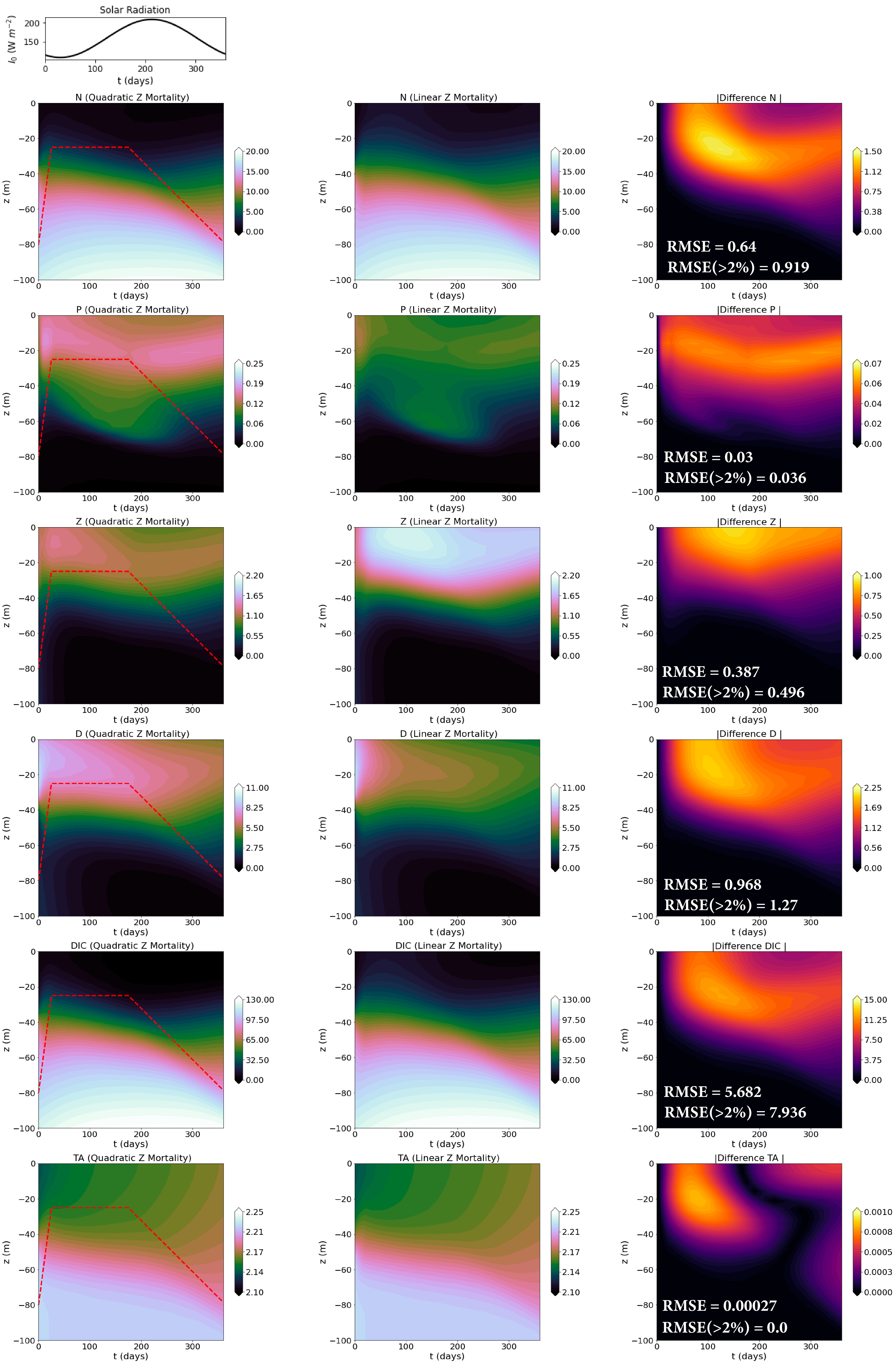}
  \caption{Solutions (in each column, concentration profiles of $N$, $P$, $Z$, $D$ in $mmol~N~m^{-3}$, $DIC$ in $mmol~C~m^{-3}$, and $TA$ in $mmol~C~kg^{-1}$, all vs.\ time in days) of the OA model used in Experiments-2a, corresponding to different functional forms for the zooplankton mortality term. 
  \textit{Left-column:} The top panel shows the yearly variation of solar radiation and the subsequent panels depict the states from the NPZD-OA model with $M_Z(Z) = \frac{m_Z}{2}(Z + Z^2)$ (ground truth), overlaid with the dynamic mixed layer depth in dashed red lines; \textit{Middle-column:} States from the NPZD-OA model with $M_Z(Z) = \frac{m_Z}{2}Z$ (low-fidelity); \textit{Right-column:} Absolute difference between the corresponding states in the left- and middle- column. For each case, we also provide the pair of time-averaged errors (see Figure \ref{fig: Exp1b_Analy_vs_LowRes_Nx50_Re1000_NBC} for description).}
  \label{fig: Exp2a_NPZD-OA_Sq_Mort_Missing}
\end{figure}

\textcolor{black}
{\emph{Training: NN architecture, data, and loss function.}}
For the \textit{g}nCM -- we only consider the Markovian term -- belonging to a linear combination of a library of popular mortality functions \cite{franks2002npz}, $\{Z, Z^2, \frac{Z^2}{1+Z}, \exp{Z}\}$. 
Compared to the true zooplankton mortality term, our library contains three superfluous or redundant terms, $Z$, $\frac{Z^2}{1+Z}$, and $\exp{Z}$, noting that the $Z$ term is already a part of the low-fidelity model and completely known.
For the  Markovian term,
we use again a simple NN with no hidden layers and linear activation in the output layer. 
Using weight constraints for the output layer, we enforce biomass conservation in the $N$, $P$, $Z$, and $D$ equations and couple with $DIC$ and $TA$ equations as in the known system (equations \ref{eq: gnCM NPZD model} and \ref{eq: gnCM OA ODE model}). Architectural details are given in table \ref{table: All architecture} and the tuned set of training hyperparameters in section \ref{sec: Hyperparameters}. 
The training data consists of the true/high-fidelity model solution sampled at time intervals of 0.1~day, until $t=$30~days, $\{\, \{B^{true}(z, T_i)\}_{B\in\{N, P, Z, D, DIC, TA\}}\}_{i=1}^M$, i.e., a $M=300$ high-fidelity solution states. 
We use an MAE-based loss function, 
$\mathcal{L} = \frac{1}{M} \sum_{i=1}^M \int_0^{D} \frac{1}{D} \sqrt{\sum_{B \in \{N, P, Z, D, DIC, TA\}} \frac{1}{\sigma_{B}}|B^{pred}(z, T_i) - B^{true}(z, T_i)|} dz$. Here, $\sigma_B$'s are hyperparameters to scale the importance of different state variables based on their magnitudes. After multiple hyperparameter tuning experiments, values of $\sigma_N = 1, ~\sigma_{P} = 0.25, ~\sigma_Z = 1, ~\sigma_D = 1, ~\sigma_{DIC} = 2, ~\sigma_{TA} = 0.1$, were found to aid in learning. 

\textcolor{black}
{\emph{Learning results.}}
In 7 repeats of the experiment with exactly the same hyperparameters, the learned models consisted of no contribution of the closure to the $N$, $P$, and $TA$ equations, while for the $Z$, $D$, and $DIC$ equations the contributions were found -- with mean and standard deviation -- to be $(-0.02996 \pm 0.00014) Z^2$, $(0.03001 \pm 0.00013) Z^2$, and $(-0.05603 \pm 0.00136) Z^2$, respectively. 
% BioOA_nODE_testcase/model_dir_case3e*
% Z: {-0.02991674, -0.0298192, -0.03011373, -0.02991661, -0.03000201,  -0.02978343, -0.03015573} -> mean = -0.02996; std = 1.405e-4
% D: { 0.02987137, 0.02981043, 0.03011962, 0.03000179, 0.02999215,  0.03017022, 0.03007047} -> 0.03001; std = 1.295e-4
% DIC: {-0.05463575, -0.05465753,  -0.05768372, -0.05557072, -0.05505633,  -0.05751297, -0.05712758} -> mean = -0.05603; std = 0.001362
For reference, the true contribution of the zooplankton quadratic mortality term to the $Z$, $D$, and $DIC$ equations are given as $-0.02998 Z^2$, $0.02998 Z^2$, and $-0.05621 Z^2$, respectively. 
%\PFJL{Need to add a sentence or two on results in Fig. 6 mid and right columns here.} 
%\AG{Added in "Comparison LF-HF:" subsection.}

\textcolor{black}
{\emph{Sensitivity.}}
Multiple experiments were done to study the effects of hyperparameters, such as batch-time, batch-size, regularization factors, etc., and the convergence to the true model was the most severely compromised when increasing batch-time and changing the loss-scaling for individual state variables.

\subsection{Experiments 2b: Ocean Acidification - Model Complexity}
%auto-ignore
\label{sec: Experiments 2b}
%\textcolor{black}{Summ.:}
In the last set of experiments, we again consider the coupled physical-biological-carbonate PDE models, however, this time we utilize the full generalized neural closure model to augment a simpler model obtained by aggregation of components and other simplifications of processes and parameterizations, such that, it becomes as accurate as the more complex model. Simultaneously, we also discriminate between the candidate functional forms of the uncertain Zooplankton mortality term in an interpretable fashion.

\textcolor{black}
{\emph{Setup: True model, data generation, and low-fidelity model.}}
% For the last set of experiments, 
%
The high-fidelity model and the data are those used in Experiments-2a (section \ref{sec: Experiments 2a}), where we model the intermediate state of detritus, thus capturing processes such as remineralization and quadratic zooplankton mortality, i.e., $M_Z(Z) = \frac{m_Z}{2}(Z + Z^2)$. 
%
%\PFJL{Add data sentence here, or we need to move all data sentences of the prior sections in the training subheaders, which is perhaps fine or better.}
%\AG{I think data is already part of the subsection "Training: NN architecture, data, and loss function." in all the exps. Thus it would make sense changing the header "Setup: True model, data, and low-fidelity model." to "Setup: high-fidelity and low-fidelity models." everywhere.}
%
The low-fidelity model is the less complex three-component NPZ model,
\begin{equation}
\label{eq: gnCM NPZ ODE}
        \begin{split}
        \frac{dN}{dt} &= -U_P + (1-\gamma) G_Z + m_P P + \frac{m_Z}{2} Z \;, \\
            \frac{dP}{dt} &= U_P - G_Z - m_P P \;, \\
            \frac{dZ}{dt} &= \gamma G_z - \frac{m_Z}{2}  Z \;, 
        \end{split}
\end{equation}
coupled with the carbonate system using fixed carbon-nitrogen ratios, $C_P$, and $C_Z$,
\begin{equation}
\label{eq: gnCM OA ODE model NPZ}
        \begin{split}
            \frac{d (DIC)}{dt} &= -C_P \frac{dP}{dt} - C_Z \frac{dZ}{dt} - \gamma_c C_P U_P \;,  \\
            \frac{d(TA)}{dt} &= -\frac{1}{\rho_w}\frac{dN}{dt} - \frac{2\gamma_c C_P U_P}{\rho_w} \,,
        \end{split}
\end{equation}
and with the 1-D diffusion-reaction PDE \ref{eq: gnCM ADR eqn}.
The goal of these experiments is to use the \textit{g}nCM to simultaneously learn the functional form of the zooplankton mortality term using the Markovian closure term, and account for the missing intermediate state of detritus through the non-Markovian closure term. 

\textcolor{black}
{\emph{Numerics.}} 
The numerical schemes used are as those of Experiments-2a in section \ref{sec: Experiments 2a}.
\textcolor{black}
{\emph{Comparison LF-HF:}}
Since the high-fidelity NPZD-OA model resolves more processes, the concentrations of $N + D$ (aggregated state), $P$, $Z$, $DIC$, and $TA$ differ significantly from the $N$, $P$, $Z$, $DIC$, and $TA$ of the low-fidelity NPZ-OA model, as shown in Figure \ref{fig: Exp2b_NPZD-OA_DistDDE}.

\begin{figure}[h!]
  \centering
  \includegraphics[width=0.775\textwidth]{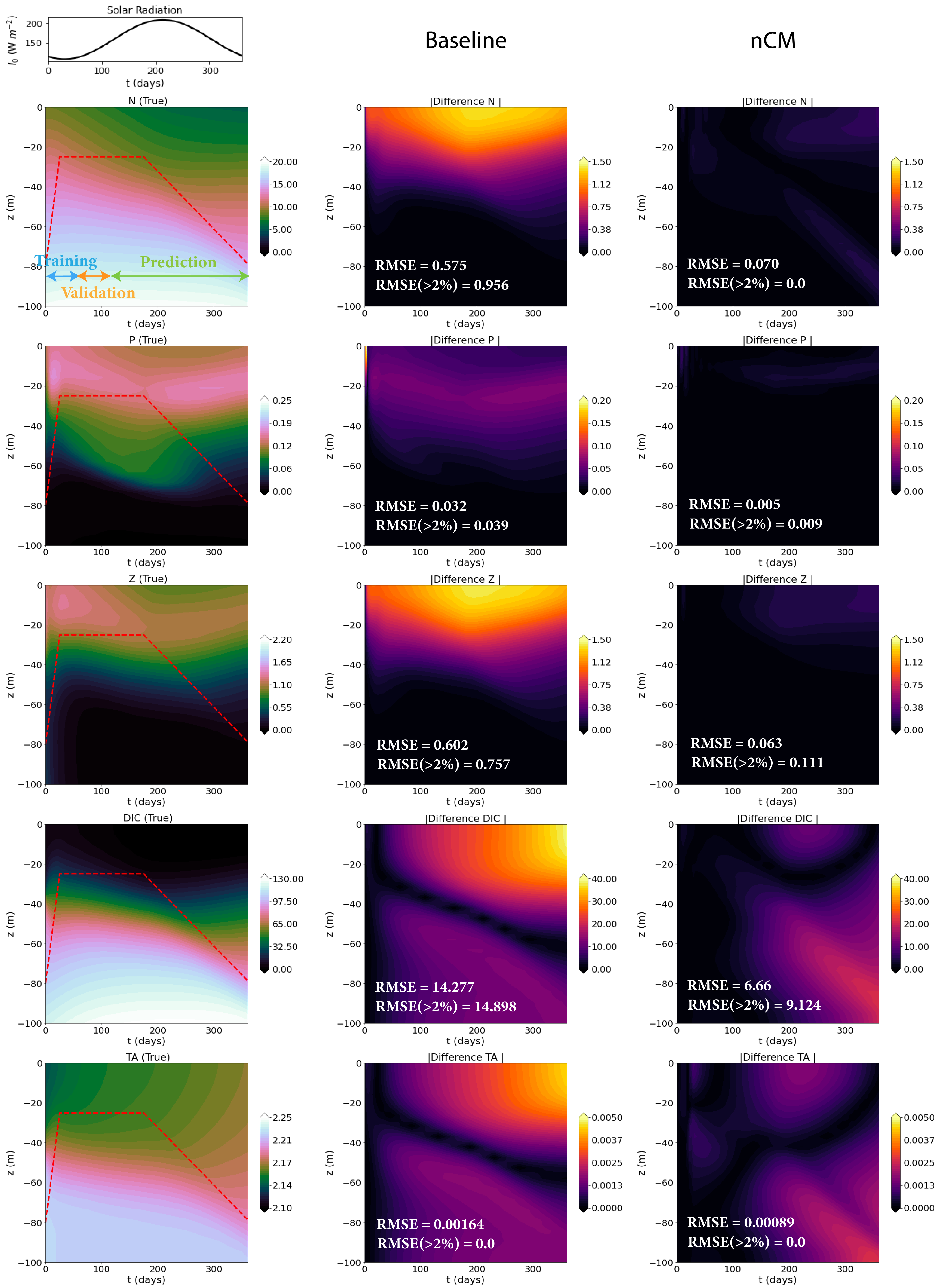}
  \caption{Comparison of the OA models used in Experiments-2b with and without closure models. The parameter values and concentration units are as in Figure \ref{fig: Exp2a_NPZD-OA_Sq_Mort_Missing}. 
  For the \textit{g}nCM, the training period is from t = 0 to 60 $days$, the validation period from t = 60 to 120 days, and the future prediction period from t = 120 to 364 days. 
  \textit{Left-column:} The top panel shows the yearly variation of solar radiation and the subsequent panels depict the aggregated states from the NPZD-OA model with $M_Z(Z) = \frac{m_Z}{2}(Z + Z^2)$ (ground truth), overlaid with the dynamic mixed layer depth in dashed red lines. \textit{Middle-column:} Absolute difference between the corresponding states from the NPZ-OA model with $M_Z(Z) = \frac{m_Z}{2}Z$ (low-fidelity) and those in the left-column (high-fidelity ground truth). 
  \textit{Right-column:} Absolute difference between the corresponding states from the low-fidelity model augmented with the learned \textit{g}nCM and the ground truth. For each case, we also provide the pair of time-averaged errors (see Figure \ref{fig: Exp1b_Analy_vs_LowRes_Nx50_Re1000_NBC} for description).
  }
  \label{fig: Exp2b_NPZD-OA_DistDDE}
\end{figure}

\textcolor{black}
{\emph{Training: NN architecture, data, and loss function.}}
Our Markovian closure consists of a linear combination of a library of popular mortality functions \cite{franks2002npz}, $\{Z,Z^2, \frac{Z^2}{1+Z},\exp Z\}$. 
Once again, compared to the true zooplankton mortality term, our library contains three redundant terms, $Z$, $\frac{Z^2}{1+Z}$, and $\exp{Z}$, where the $Z$ term is already a part of the low-fidelity model and completely known.
Additionally, we use a deep-NN for the non-Markovian closure term, with $N(z, t)$, $P(z, t)$, $Z(z, t)$, and $I(z, t)$ as the input; the inclusion of the photosynthetically active radiation, $I(z, t)$, makes this closure term non-autonomous. 
The architecture for the fully-connected deep-NN used in the non-Markovian closure term is provided in table \ref{table: All architecture}, and it consists of two hidden-layers with the non-linear \textit{swish} activation.  
We do not include the states $DIC(z, t)$ and $TA(z, t)$ among the inputs in order to preserve one-way coupling between the biological and carbonate system. 
Along with this, biomass conservation and coupling of the carbonate system by nitrogen conversion (as in equations \ref{eq: gnCM NPZ ODE} and \ref{eq: gnCM OA ODE model NPZ}) is maintained in the non-Markovian closure terms by manipulating the channels of the output layer. 
On the other hand, in the Markovian layer, these constraints are imposed by constraining the weights of the output layer. To help with learning, we further impose the condition that the contribution of the Markovian closure term to the $P$ equation is exactly equal to zero. See table \ref{table: All architecture} for implementational details of these constraints.

The training data consists of solving the NPZD-OA model with $M(Z) = \frac{m_Z}{2}(Z + Z^2)$, and the solution sampled at time intervals of 0.1~day until $t=  60~days$, $\{\{B^{true}(z, T_i)\}_{B\in\{N+D, P, Z, DIC, TA\}}\}_{i=1}^M$, i.e., $M=600$ high-fidelity solution states at different times.
%\PFJL{As mentioned before, give the value of M here, How much data can be provided at most?}
Performance of the learned model in the validation interval of 60~days\;$\leq t \leq$\;120~days is used to tune the hyperparameters, provided in section \ref{sec: Hyperparameters}. We again use a MAE based loss function, $\mathcal{L} = \frac{1}{M} \sum_{i=1}^M \int_0^{D} \frac{1}{D} \sqrt{\sum_{B \in \{N, P, Z, DIC, TA\}} \frac{1}{\sigma_{B}}|B^{pred}(z, T_i) - B^{true}(z, T_i)|} dz$, with $\sigma_N = 1, ~\sigma_{P} = 0.25, ~\sigma_Z = 1, ~\sigma_{DIC} = 2, ~\sigma_{TA} = 0.1$ (similar to those used in experiments-2a). A time delay of $\tau = 2.5~days$ was used for the non-Markovian closure term based on the optimal delay value study performed in \cite{gupta_lermusiaux_PRSA2021}.

\textcolor{black}
{\emph{Learning results.}}
In 9 repeats of the experiment with exactly the same set of hyperparameters, the mean and standard deviation of the learned contribution of the Markovian closure term to the $Z$ equation is given by, $(-0.03000 \pm 0.00067) Z^2$. 
% NPZOA_nDistDDE_testcase/model_dir_case11d*
% Z: {-0.02973846, -0.03009911, -0.03091634, -0.0294171, -0.02942759,  -0.02931909,  -0.02988794, -0.03108947, -0.03007241} -> mean = -0.03; std = 6.377e-4
% DIC: {0.16727881, 0.1693075,  0.17390439, 0.16547118, 0.16553019,  0.16491987, 0.16811965, 0.17487828, 0.1691573} -> mean = 0.1687; std = 0.003587
For reference, the true contribution of the quadratic mortality term to the $Z$ equation is $-0.02998 Z^2$. Due to the weight constraints, the contribution of the Markovian closure term to other equations is exactly zero. We evaluate the performance of the learned neural closure model for long predictions, spanning over 1 year ($365~days$). 
The comparison with true/high-fidelity data for one of the experiments is provided in Figure \ref{fig: Exp2b_NPZD-OA_DistDDE}. 
Overall, the learned closure keeps the errors low throughout the 1-year time period, apart from a slight increase observed for the OA states after $\sim 200~days$. 

\textcolor{black}
{\emph{Sensitivity.}}
Multiple experiments were done to study the effects of hyperparameters, such as batch-time, batch-size, regularization factors, etc., and their effects were similar to those observed in previous experiments. 
However, when using larger neural network architectures for the non-Markovian term, this led to high variability in the learned coefficients of the Markovian term on repeats of the experiments with the same set of hyperparameters. This is probably because of the increased expressive power of the non-Markovian term, which overshadows the significance of the learned Markovian term.

% \subsection{Computational Advantages}
% \input{Sections/Sec_gnCM_computational_advantage}

\subsection{Remarks and Discussion}

\paragraph{\textit{Computational advantages.}}
%auto-ignore
In \cite{gupta_lermusiaux_PRSA2021}, through a flop-count analysis, we proved that the additional computational cost due to the presence of neural closure models is of similar or lower complexity than the existing low-fidelity model. However, in our current generalized framework, we have additional computational advantages. 
First, the size of the neural network architecture is completely independent of the number of discretized state variables and only dictated by the number of local features to be used as inputs to the \textit{g}nCM terms. 
Second, as the same neural networks are applied locally at every grid point, it is directly possible to use batches of the size of the number of grid points. 
It has been reported that larger batch sizes could lead to performance speed-ups in forward pass through neural networks during the inference stage \cite{kochura2019batch}.
Estimating the leading flop-count order for training is non-trivial due to the presence of a number of operations ranging from time-integration of the forward model and adjoint PDEs; automatic differentiation through the neural networks; creation and use of interpolation functions; the integral to compute the final derivatives; the gradient descent step, etc. All these operations lead to training costs that are non-negligible. However, the generalizability and interpretability of our learned \textit{g}nCMs over boundary conditions, initial conditions, domain, problem-specific parameters, etc., help justify the one-time training cost. 

\paragraph{\textit{Lack of prior knowledge.}} 
As showcased in the prior experiments and summarized in the corresponding sensitivity studies,
the lack of prior knowledge about the missing dynamics could manifest in many different ways. 
This includes no known low-fidelity model, dynamics of the known low-fidelity model very different from the high-fidelity model/data, no knowledge of potential candidate terms to create input function libraries, or even no information on the most relevant state variables themselves. 
To allow compensation for this lack of prior knowledge,
our \textit{g}nCM framework is derived and implemented for any deep-neural-network (DNN) architectures for both Markovian and non-Markovian closure terms 
(Figure~\ref{fig: framework overview}). 
Our flexible modeling framework provides full autonomy for the design of the unknown closure terms such as using linear-, shallow-, or deep-NNs, selecting the span of the function libraries, and using either or both Markovian and non-Markovian closure terms.
All these decisions are made by the subject matter expert/user depending on the problem at hand. 
For example, in all our experiments, fully-connected deep-NNs were utilized for the non-Markovian closure terms, because in general, no prior knowledge is available for the same. Further, our framework could be extended to allow for the adaptive increase of the input function library, for example using the algorithm proposed in \cite{kulkarni_et_al_DDDAS2020}.

\paragraph{\textit{Non-Markovian term.}}
In the current derivation of the \textit{g}nCM framework, due to the mathematical constructs, the non-Markovian term does not account for the possibility of memory decay contribution of the $\mathcal{D}_{NN}(\bu)$ function under the integral w.r.t.\, $t-s$ in equation~\ref{eq: low-fid with closure terms}. However, memory decay or other variations can be a desired property for some problems. To allow for this, one can split the integral in equation~\ref{eq: low-fid with closure terms} into contiguous parts and multiply each of them with different weights.
An alternate option is to consider discrete delays utilizing recurrent neural networks as for the discrete-nDDEs in \cite{gupta_lermusiaux_PRSA2021} and so implicitly incorporate the desired memory decay.
In general, the need for the non-Markovian closure term for a given problem should be determined by the subject matter expert.  
However, in many cases, we anticipate the need for non-Markovian closure term to be imperative, especially when the high-fidelity model/data accounts for intermediate state variables not modeled in the known low-fidelity model, as in our Experiments-2b (section~\ref{sec: Experiments 2b}). 
Finally, it is also desirable to allow learning an adaptive optimal delay ($\tau$ in equation~\ref{eq: low-fid with closure terms}), instead of treating it as a hyperparameter. For such a possibility, we refer to Appendix~E in \cite{gupta_PhDThesis2022} where we derive the theory for learning the optimal delay for the nCM framework (\cite{gupta_lermusiaux_PRSA2021}; DDE counterpart for \textit{g}nCM).

%%%%%%%%%%%%%%%%%%%%%%%%%%%%%%%%%%%%%%%%%%%%%%%%%%%%%%%%%
%%%%%%%%%%%%%%%%%%%%%%%%%%%%%%%%%%%%%%%%%%%%%%%%%%%%%%%%%

\section{Conclusions}
\label{sec: conclusions}
%auto-ignore

In the present study, we develop neural closure models 
%\cite{gupta_lermusiaux_PRSA2021}
that are %readily 
generalizable over computational grid resolution, boundary and initial conditions, domain geometries, and problem parameters, and also provide interpretability. These generalized neural closure models (\textit{g}nCMs) are based on neural partial delay differential equations (nPDDEs) that augment existing/low-fidelity models in their PDE forms with both Markovian and non-Markovian closures parameterized with deep-NNs. 
The melding in the continuous spatiotemporal space is then followed by numerical discretization. This ensures that the burden of generalization, along with computing the relevant spatial derivatives, is carried by the numerical schemes, and not by the learned NNs. 
The space-time continuous form of the \textit{g}nCMs also makes it easy to interpret the learned closures. 
For efficient training, we derive the adjoint PDEs in the continuous form and discretize them with adaptive time-integration schemes and employ interpolation functions constructed during forward integration
%checkpointing, and IRDM
to increase numerical stability and accuracy. 
%\AG{We have not implemented checkpointing/IRDM.}
%\PFJL{Yes, but you use interpolation functions. We can discuss this so I understand and then I can fix my text.} \AG{Sounds good.}
%
This enables implementation across differentiable and non-differentiable computational physics codes, and different machine learning frameworks, all while being agnostic to the numerical methods. It further removes any requirements on the availability of regularly spaced training data in both space and time, and also accounts for errors in the time-evolution of the states in the presence of NNs during training.
Finally, all our derivations and implementations consider deep-NN architectures for both Markovian and non-Markovian terms, thus automatically encompassing linear- and shallow-NNs, and providing the user or subject-matter-expert with the flexibility of choosing the architectural complexity in accord with prior knowledge.

Through a series of four sets of experiments, we demonstrate the interpretability and generalizability of our learned \textit{g}nCMs. Our first two sets of simulation experiments are based on advecting nonlinear waves and shocks governed by the KdV-Burgers and classic Burgers PDEs, where the low-fidelity models are either missing terms or contain errors due to unresolved subgrid-scale processes. 
When presented with a function library containing terms of spatial derivatives of different orders and their combinations, grid-resolution, and the Reynolds number as inputs to the closure terms, our learned \textit{g}nCMs 
eliminate redundant terms and
discover missing physics, leading truncation error terms, and a correction to the nonlinear advection, all in an interpretable fashion.
The correction to the nonlinear advection term, despite being absent from the input function library, is still accounted for and learned indirectly.
Further, by analyzing the deep-NN weights, we also notice the learned closure terms to be independent of the Reynolds number. We find that training on data corresponding to just $3-4$ combinations of a number of grid points and Reynolds number is sufficient to ensure that the learned closures are generalizable over large ranges of grid resolutions and Reynolds numbers, initial and boundary conditions, and also outperform the popular Smagorinsky closure model. 
Our last two sets of experiments are based on one-dimensional, non-autonomous ocean acidification PDE models, that couple physical, biological, and carbonate states, processes, and interactions. In these experiments, the low-fidelity models have uncertainty in the functional form of certain biological processes and lack complexity due to a missing intermediate state variable. 
The learned \textit{g}nCMs
simultaneously discriminate between candidate functional forms of the uncertain Zooplankton mortality term with the Markovian part of the closure, and account for the missing intermediate state and processes with the non-Markovian part. 
%
% Our framework is shown to provide the flexibility of using deep-NNs for both Markovian and non-Markovian closure terms, which can help compensate for the lack of prior knowledge about the functional form of the missing dynamics.
%
In terms of computational advantage, our new framework naturally lends itself to batching across computational grid points during the forward pass through the NNs in the closure terms, thus leading to potential performance speed-ups.

The \textit{g}nCMs allow learning both Markovian and non-Markovian closure parameterization with deep-NNs at the PDE level, thus addressing the issues of generalizability and interpretability that are often the bottleneck when it comes to using machine learning for computational science and engineering problems. The generalizability and interpretability properties also make it easier to justify the often computationally expensive training stage, thus enabling wider adoption.

\vspace{0.75cm}

% \ethics{Insert ethics text here.}

\noindent\textbf{Data Accessibility. }{The codes and data used in this work are available in the GitHub repository: \url{https://github.com/mit-mseas/generalized_nCMs.git}}

\vspace{0.25cm}

\noindent\textbf{Author's Contributions. }{A.G.\ conceived the idea of extending the existing neural closure models using neural partial delay differential equations; augmentation of closure terms in the continuous spatio-temporal space followed with numerical discretization for generalizability and interpretability; derived the adjoint PDE; implemented the neural network architectures and the simulation experiments; interpreted the computational results; and wrote a first draft of the manuscript. P.F.J.L.\ supervised the work; provided ideas; interpreted the results; and edited and wrote significant parts of the manuscript.}

\vspace{0.25cm}

\noindent\textbf{Competing Interests. }{We declare we have no competing interests.}

\vspace{0.25cm}

\noindent\textbf{Funding. }{We are grateful to the Office of Naval Research for partial support under
grant N00014-20-1-2023 (MURI ML-SCOPE) to the
Massachusetts Institute of Technology.}

\vspace{0.25cm}

\noindent\textbf{Acknowledgement. }{We are grateful to the members of our MSEAS group for their collaboration and insights, especially Mr.\ Aman Jalan. We thank the members of our MURI ML-SCOPE research team for many useful discussions. We also thank the three anonymous reviewers for their useful comments.
}

% \disclaimer{Insert disclaimer text here.}

%%%%%%%%% Add appendix here %%%%%%%%%%%%%%%%%%%%%

%%%%%%%%%% Insert bibliography here %%%%%%%%%%%%%%

% \bibliographystyle{unsrtnat}
\bibliographystyle{abbrvnat}
\bibliography{mseas,ml,biology,additional}
% \printbibliography
\makeatletter\@input{msxx.tex}\makeatother
\end{document}

\date{\today}
\maketitle

\section{Adjoint Equations for Neural Partial Delay Differential Equations}
\label{sec: gnCM Adjoint Equations in supp info}
%auto-ignore
In this section, we provide a detailed derivation of adjoint equations for neural partial delay differential equations (nPDDEs). This derivation is inspired by the adjoint equation derivation for a general PDE by Li and Petzold, 2004 \cite{li2004adjoint} and Cao et al., 2002 \cite{cao2002adjoint}. Our nPDDE is of the form,
\begin{equation}
    \begin{split}
        \frac{\partial u(x, t)}{\partial t} =& \mathcal{F}_{NN}\left(u(x, t), \frac{\partial u(x, t)}{\partial x}, \frac{\partial^2 u(x, t)}{\partial x^2}, ..., \frac{\partial^d u(x, t)}{\partial x^d}, x, t; \phi \right) \\
        & + \int_{t-\tau}^t \mathcal{D}_{NN}\left(u(x, s), \frac{\partial u(x, s)}{\partial x}, \frac{\partial^2 u(x, s)}{\partial x^2}, ..., \frac{\partial^d u(x, s)}{\partial x^d}, x, s; \theta\right)ds \,, \\
        & \hspace{0.6\textwidth} x\in \Omega, \; t \geq 0, \\
        u(x, t) &= h(x, t), \; -\tau \leq t \leq 0 \quad \text{and} \quad \mathcal{B}(u(x, t)) = g(x, t), \; x \in \partial \Omega, \; t \geq 0 \;.
    \end{split}
    \label{eq(SI): nPDDE in original form}
\end{equation}
As compared to PDEs, PDDEs require the specification of a history function ($h(x, t)$) for the initial conditions. $\mathcal{F}_{NN}(\bu; \phi)$ and $\mathcal{D}_{NN}(\bu; \theta)$ are two neural networks (NNs) parameterized by $\phi$ and $\theta$, respectively. We consider the presence of an arbitrary number of spatial derivatives, with the highest order defined by $d \in \mathbb{Z}^+$. We can rewrite the above equation \ref{eq(SI): nPDDE in original form} as an equivalent system of coupled PDDEs with discrete delays,
\begin{equation}
    \begin{split}
        \frac{\partial u(x, t)}{\partial t} =& \mathcal{F}_{NN}\left(u(x, t), \frac{\partial u(x, t)}{\partial x}, \frac{\partial^2 u(x, t)}{\partial x^2}, ..., \frac{\partial^d u(x, t)}{\partial x^d}, x, t; \phi \right) + y(x, t) \;,  \\
        & \hspace{0.6\textwidth} x\in \Omega, \; t \geq 0, \\
        \frac{\partial y(x, t)}{\partial t} =& \mathcal{D}_{NN}\left(u(x, t), \frac{\partial u(x, t)}{\partial x}, \frac{\partial^2 u(x, t)}{\partial x^2}, ..., \frac{\partial^d u(x, t)}{\partial x^d}, x, t; \theta\right) \\
        & - \mathcal{D}_{NN}\left(u(x, t-\tau), \frac{\partial u(x, t-\tau)}{\partial x}, \frac{\partial^2 u(x, t-\tau)}{\partial x^2}, ..., \frac{\partial^d u(x, t-\tau)}{\partial x^d}, x, t-\tau; \theta\right) \;,  \\
        & \hspace{0.6\textwidth} x\in \Omega, \; t \geq 0,\\
        u(x, t) =& h(x, t), -\tau \leq t \leq 0 \quad \text{and} \quad\mathcal{B}(u(x, t)) = g(x, t), x \in \partial \Omega, t \geq 0 \\
        y(x, 0) =& \int_{-\tau}^0 \mathcal{D}_{NN}\left(h(x, s), \frac{\partial h(x, s)}{\partial x}, \frac{\partial^2 h(x, s)}{\partial x^2}, ..., \frac{\partial^d h(x, s)}{\partial x^d} x, s; \theta\right)ds \,.
    \end{split}
\end{equation}
Let us assume that high-fidelity data is available at M discrete times, $T_1 < ...<T_M \leq T$, and at $N(T_i)$ spatial locations ($x_{k}^{T_i} \in \Omega, \forall k \in {1, ..., N(T_i)}$) for each of the times. We define the scalar loss function as $L = \frac{1}{M} \sum_{i=1}^{M} \frac{1}{N(T_i)}\sum_{k=1}^{N(T_i)} l(u(x^{T_i}_k, T_i)) \equiv \int_0^T \frac{1}{M} \sum_{i=1}^{M} \int_{\Omega} \frac{1}{N(T_i)}\sum_{k=1}^{N(T_i)} l(u(x, t))\delta(x - x^{T_i}_k) \delta(t - T_i)dxdt \allowbreak \equiv \int_0^T \frac{1}{M} \sum_{i=1}^{M} \frac{1}{|\Omega|} \int_{\Omega} \hat{l}(u(x, t))\delta(t - T_i) dxdt$, where $l(\bu)$ are scalar loss functions such as mean-squared-error (MSE), and $\delta(\bu)$ is the Kronecker delta function. In order to derive the adjoint PDEs, we start with the Lagrangian corresponding to the above system,
\begin{equation}
    \begin{split}
        \mathbb{L} =& L(u(x, t)) + \iw \lt(x, t) \left(\p_t u(x, t) - \mathcal{F}_{NN}(\bu, t; \phi) - y(x, t)\right) dxdt \\
        & + \iw \mt(x, t) \left(\p_t y(x, t) - \mathcal{D}_{NN}(\bu, t;\theta) + \mathcal{D}_{NN}(\bu, t-\tau; \theta)\right) dxdt \\
        & + \int_{\Omega} \alpha^T(x) \left(y(x, 0) - \int_{-\tau}^0 \mathcal{D}_{NN}(h(x, t), \p_x h(x, t), \p_{x^2} h(x, t), ..., \p_{x^d} h(x, t), x, t; \theta)dt\right)dx \,,
    \end{split}
    \label{eq(SI): lagrangian}
\end{equation}
where $\lambda(x, t)$, $\mu(x, t)$ and $\alpha(x)$ are the Lagrangian variables. We then take the derivative of the Lagrangian (equation \ref{eq(SI): lagrangian}) w.r.t. $\theta$ (for brevity we denote, $\partial / \partial (\bu) \equiv \partial_{(\bu)}$, and $d / d (\bu) \equiv d_{(\bu)}$),
% MB: Not sure if the brevity thing here is similar to what I saw before in the main chapter. I think the dot should be a subscript for the short hand notation.
\begin{equation}
    \begin{split}
        \dt\mathbb{L} =& \iw  \frac{1}{M}\frac{1}{|\Omega|} \sum_{i=1}^{M} \p_{u(x, t)}\hat{l}(u(x, t))\delta(t - T_i) \dt u(x, t) dxdt + \iw \lt(x, t)\left(\p_t \dt u(x, t) \right.\\
        & \left. - \p_{u(x, t)}\mathcal{F}_{NN}(\bu, t; \phi) \dt u(x, t)  - \p_{\p_x u(x, t)} \mathcal{F}_{NN}(\bu, t; \phi) \dt\p_x u(x, t)  \right.\\
        & \left. - \p_{\p_{xx}u(x, t)} \mathcal{F}_{NN}(\bu, t; \phi) \dt \p_{xx}u(x, t) - \dt y(x, t)\right) dxdt 
        + \iw \mt(x, t) \left(\p_t \dt y(x, t) \right.\\
        & \left. - \p_{u(x, t)}\mathcal{D}_{NN}(\bu, t;\theta) \dt u(x, t) - \p_{\p_x u(x, t)} \mathcal{D}_{NN}(\bu, t; \theta) \dt\p_x u(x, t) \right.\\
        & \left. - \p_{\p_{xx} u(x, t)} \mathcal{D}_{NN}(\bu, t; \theta) \dt \p_{xx}u(x, t) \right.\\
        & \left. - \partial_{\theta} \mathcal{D}_{NN}(\bu, t; \theta)
        + \p_{u(x, t)}\mathcal{D}_{NN}(\bu, t-\tau;\theta) \dt u(x, t-\tau)  \right.\\
        & \left. + \p_{\p_x u(x, t-\tau)} \mathcal{D}_{NN}(\bu, t-\tau; \theta) \dt\p_x u(x, t-\tau) \right.\\
        & \left. + \p_{\p_{xx} u(x, t-\tau)} \mathcal{D}_{NN}(\bu, t-\tau; \theta) \dt \p_{xx}u(x, t-\tau) + \partial_{\theta} \mathcal{D}_{NN}(\bu, t - \tau; \theta)\right) dxdt \\
        & + \int_{\Omega} \alpha^T(x) \left( \dt y(x, 0) - \int_{-\tau}^0 \pth G_{NN}(h(x, t), \p_x h(x, t), \p_{xx} h(x, t), x, t; \theta)dt \right)dx \,.
    \end{split}
\end{equation}
Using integration-by-parts, we get,
\begin{equation}
    \begin{split}
        \iw \lt(x, t) \p_t\dt u(x, t) dxdt =& \int_{\Omega} [\lt(x, t) \dt u(x, t)] \bigg|_0^T - \iw \p_t \lt(x, t) \dt u(x, t) dxdt \,,
    \end{split}
\end{equation}
\begin{equation}
    \begin{split}
        \iw \lt(x, t) \p_{\p_x u(x, t)} & \mathcal{F}_{NN}(\bu, t) \p_x\dt u(x, t) dxdt \\
        =&\ib \lt(x, t)\p_{\p_x u(x, t)} \mathcal{F}_{NN}(\bu, t)\dt u(x, t)dxdt \\
        & - \iw \p_x \left(\lt(x, t) \p_{\p_x u(x, t)} \mathcal{F}_{NN}(\bu, t) \right) \dt u(x, t)dxdt \,,
    \end{split}
\end{equation}
\begin{equation}
    \begin{split}
        \iw \lt(x, t) \p_{\p_{xx} u(x, t)} & \mathcal{F}_{NN}(\bu, t) \p_{xx}\dt u(x, t)dxdt  \\
         = & \ib \lt(x, t)\p_{\p_{xx} u(x, t)} \mathcal{F}_{NN}(\bu, t) \p_x\dt u(x, t)dxdt \\
        & - \iw \p_{x} \left(\lt(x, t) \p_{\p_{xx} u(x, t)} \mathcal{F}_{NN}(\bu, t) \right) \p_x \dt u(x, t)dxdt \\
        = & \ib \lt(x, t)\p_{\p_{xx} u(x, t)} \mathcal{F}_{NN}(\bu, t) \p_x \dt u(x, t)dxdt \\
        & - \left( \ib \p_x \left(\lt(x, t) \p_{\p_{xx} u(x, t)} \mathcal{F}_{NN}(\bu, t)\right) \dt u(x, t)dxdt \right.\\
        & \left. - \iw \p_{xx} \left( \lt(x, t) \p_{\p_{xx} u(x, t)} \mathcal{F}_{NN}(\bu, t) \right) \dt u(x, t)dxdt \right) \,,
    \end{split}
\end{equation}
\begin{equation}
    \begin{split}
            \iw \mt(x, t) \p_{u(x, t-\tau)} &\mathcal{D}_{NN}(\bu, t-\tau) \dt u(x, t-\tau) dxdt \\
            =& \int_{-\tau}^{T-\tau}\int_{\Omega} \mt(x, t+\tau) \p_{u(x, t)} \mathcal{D}_{NN}(\bu, t) \dt u(x, t)  dxdt \\
            =& \int_{0}^{T}\int_{\Omega} \mt(x, t+\tau) \p_{u(x, t)} \mathcal{D}_{NN}(\bu, t) \dt u(x, t) dxdt\\
            & - \int_{T-\tau}^{T}\int_{\Omega} \mt(x, t+\tau) \p_{u(x, t)} \mathcal{D}_{NN}(\bu, t) \dt u(x, t) dxdt \\
            & + \int_{-\tau}^{0}\int_{\Omega} \mt(x, t+\tau) \p_{u(x, t)} \mathcal{D}_{NN}(\bu, t) \dt u(x, t) dxdt \,.
    \end{split}
\end{equation}
Using the fact that $\mt(x, t) = 0, \forall t \geq T$ and $\dt u(x, t) = 0, -\tau \leq t\leq 0$, we get,
\begin{equation}
    \begin{split}
            \iw \mt(x, t) \p_{u(x, t-\tau)} &\mathcal{D}_{NN}(\bu, t-\tau; \theta) \dt u(x, t-\tau) dxdt \\
            =& \int_{0}^{T}\int_{\Omega} \mt(x, t+\tau) \p_{u(x, t)} \mathcal{D}_{NN}(\bu, t; \theta) \dt u(x, t) dxdt \,.
    \end{split}
\end{equation}
Similarly, we also get,
\begin{equation}
    \begin{split}
        \iw \mt(x, t) \p_{\p_x u(x, t - \tau)} &\mathcal{D}_{NN}(\bu, t-\tau; \theta) \p_x\dt u(x, t-\tau) dxdt \\
        =&\ib \mt(x, t + \tau)\p_{\p_x u(x, t)} \mathcal{D}_{NN}(\bu, t; \theta)\dt u(x, t)dxdt \\
        & - \iw \p_x \left(\mt(x, t+\tau) \p_{\p_x u(x, t)} \mathcal{D}_{NN}(\bu, t; \theta) \right) \dt u(x, t)dxdt \,,
    \end{split}
\end{equation}
\begin{equation}
    \begin{split}
        \iw \mt(x, t) \p_{\p_{xx} u(x, t-\tau)} & \mathcal{D}_{NN}(\bu, t-\tau; \theta) \p_{xx}\dt u(x, t-\tau)dxdt \\
        =& \ib \mt(x, t+\tau)\p_{\p_{xx} u(x, t)} \mathcal{D}_{NN}(\bu, t; \theta) \p_x \dt u(x, t)dxdt \\
        & - \left( \ib \p_x \left(\mt(x, t+\tau) \p_{\p_{xx} u(x, t)} \mathcal{D}_{NN}(\bu, t; \theta)\right) \dt u(x, t)dxdt \right.\\
        & \left. - \iw \p_{xx} \left( \mt(x, t+\tau) \p_{\p_{xx} u(x, t)} \mathcal{D}_{NN}(\bu, t; \theta) \right) \dt u(x, t)dxdt \right) \,.
    \end{split}
\end{equation}
Plugging everything in yields,
\begingroup
\allowdisplaybreaks
\begin{align}
    % \begin{split}
        \dt\mathbb{L} =& \iw \frac{1}{M}\frac{1}{|\Omega|} \sum_{i=1}^{M} \p_{u(x, t)}\hat{l}(u(x, t))\delta(t - T_i) \dt u(x, t)dxdt \nonumber\\
        & - \iw \p_t \lt(x, t) \dt u(x, t) dxdt \nonumber\\
        & - \iw \lt(x, t)\p_{u(x, t)}\mathcal{F}_{NN}(\bu, t) \dt u(x, t) dxdt \nonumber\\
         & - \ib \lt(x, t)\p_{\p_x u(x, t)} \mathcal{F}_{NN}(\bu, t)\dt u(x, t)dxdt \nonumber\\
         & + \iw \p_x \left(\lt(x, t) \p_{\p_x u(x, t)} \mathcal{F}_{NN}(\bu, t) \right) \dt u(x, t)dxdt \nonumber\\
         &- \ib \lt(x, t)\p_{\p_{xx} u(x, t)} \mathcal{F}_{NN}(\bu, t) \p_x \dt u(x, t)dxdt  \nonumber\\
         & +  \ib \p_x \left(\lt(x, t) \p_{\p_{xx} u(x, t)} \mathcal{F}_{NN}(\bu, t)\right) \dt u(x, t)dxdt \nonumber\\
        &  - \iw \p_{xx} \left( \lt(x, t) \p_{\p_{xx} u(x, t)} \mathcal{F}_{NN}(\bu, t) \right) \dt u(x, t)dxdt  \nonumber\\
        & - \iw \lt(x, t) \dt y(x, t) dxdt \nonumber\\
        & - \int_{\Omega} \mt(x, 0) \dt y(x, 0) dx - \iw \p_t \mt (x, t) \dt y(x, t) dxdt \nonumber\\
        & - \iw \mt(x, t) \p_{u(x, t)} \mathcal{D}_{NN}(\bu, t; \theta) \dt u(x, t) dxdt \nonumber\\
        & - \ib \mt(x, t)\p_{\p_x u(x, t)} \mathcal{D}_{NN}(\bu, t; \theta)\dt u(x, t)dxdt \nonumber\\
        & + \iw \p_x \left(\mt(x, t) \p_{\p_x u(x, t)} \mathcal{D}_{NN}(\bu, t; \theta) \right) \dt u(x, t)dxdt \nonumber\\
        & - \ib \mt(x, t)\p_{\p_{xx} u(x, t)} \mathcal{D}_{NN}(\bu, t; \theta) \p_x \dt u(x, t)dxdt  \nonumber\\
        & + \ib \p_x \left(\mt(x, t) \p_{\p_{xx} u(x, t)} \mathcal{D}_{NN}(\bu, t; \theta)\right) \dt u(x, t)dxdt \nonumber\\
        &  - \iw \p_{xx} \left( \mt(x, t) \p_{\p_{xx} u(x, t)} \mathcal{D}_{NN}(\bu, t; \theta) \right) \dt u(x, t)dxdt  \nonumber\\
        & - \iw \mt(x, t) \pth \mathcal{D}_{NN}(\bu, t; \theta) dxdt \nonumber\\
        & + \iw \mt(x, t+\tau) \p_{u(x, t)} \mathcal{D}_{NN}(\bu, t; \theta) \dt u(x, t) dxdt \nonumber\\
        & + \ib \mt(x, t+\tau)\p_{\p_x u(x, t)} \mathcal{D}_{NN}(\bu, t; \theta)\dt u(x, t)dxdt \nonumber\\
        & - \iw \p_x \left(\mt(x, t+\tau) \p_{\p_x u(x, t)} \mathcal{D}_{NN}(\bu, t; \theta) \right) \dt u(x, t)dxdt \nonumber\\
        & + \ib \mt(x, t+\tau)\p_{\p_{xx} u(x, t)} \mathcal{D}_{NN}(\bu, t; \theta) \p_x \dt u(x, t)dxdt  \nonumber\\
        & - \ib \p_x \left(\mt(x, t+\tau) \p_{\p_{xx} u(x, t)} \mathcal{D}_{NN}(\bu, t; \theta)\right) \dt u(x, t)dxdt \nonumber\\
        &  + \iw \p_{xx} \left( \mt(x, t+\tau) \p_{\p_{xx} u(x, t)} \mathcal{D}_{NN}(\bu, t; \theta) \right) \dt u(x, t)dxdt  \nonumber\\
        & + \iw \mt(x, t) \pth \mathcal{D}_{NN}(\bu, t-\tau; \theta) dxdt \nonumber\\
        & + \int_{\Omega} \alpha^T(x)  \dt y(x, 0) \nonumber\\
        & - \int_{\Omega} \alpha^T(x) \int_{-\tau}^0 \pth \mathcal{D}_{NN}(h(x, t), \p_x h(x, t), \p_{xx} h(x, t), x, t; \theta)dt dx \,. \qquad\qquad\qquad 
    % \end{split}
\end{align}
\endgroup
Collecting all the terms with $\int_{\Omega}$, $\dt u(x, t)$, and $\dt y(x, t)$, we get the following adjoint PDEs,
\begin{equation}
\label{eq(SI): adjoint PDEs}
    \begin{split}
        0 &= \frac{1}{M}\frac{1}{|\Omega|} \sum_{k=1}^M \p_{u(x, t)} \hat{l}(u(x, t)) \delta(t - T_k) \\
        & - \p_t \lt(x, t) -  \lt(x, t)\p_{u(x, t)}\mathcal{F}_{NN}(\bu, t) 
         + \sum_{i=1}^d (-1)^{i+1} \p_{x^i} \left(\lt(x, t) \p_{\p_{x^i} u(x, t)} \mathcal{F}_{NN}(\bu, t) \right)
        \\
        & - \mt(x, t) \p_{u(x, t)} \mathcal{D}_{NN}(\bu, t; \theta)
        + \sum_{i=1}^d (-1)^{i+1} \p_{x^i} \left(\mt(x, t) \p_{\p_{x^i} u(x, t)} \mathcal{D}_{NN}(\bu, t; \theta) \right)
         \\
        & + \mt(x, t+\tau) \p_{u(x, t)} \mathcal{D}_{NN}(\bu, t; \theta)
        - \sum_{i=1}^d (-1)^{i+1} \p_{x^i} \left(\mt(x, t+\tau) \p_{\p_{x^i} u(x, t)} \mathcal{D}_{NN}(\bu, t; \theta) \right) \;, \\
        & \hspace{0.6\textwidth} x \in \Omega \;, \; t \in [0, T) \,,
         \\
        0 & = -\lt(x, t) - \p_t \mt (x, t) \;, \quad x \in \Omega \;, \; t \in [0, T) \,,
    \end{split}
\end{equation}
with initial conditions, $\lambda(x, t) =\mu (x, t) = 0, \; t \geq T$. The boundary conditions are derived based on that of the forward PDE such that they satisfy,
\begin{equation}
\label{eq(SI): adjoint PDEs BCs}
    \begin{split}
        0 &= \sum_{i = 0}^d \sum_{j=0}^{d-i-1} (-1)^{j+1} \p_{x^{j}} \left(\lt(x, t)\p_{\p_{x^{j+i+1}} u(x, t)} \mathcal{F}_{NN}(\bu, t)\right)\dt \p_{x^{i}}u(x, t) 
        \\
        & +\sum_{i = 0}^d\sum_{j=0}^{d-i-1} (-1)^{j+1} \p_{x^{j}} \left(\mt(x, t)\p_{\p_{x^{j+i+1}} u(x, t)} \mathcal{D}_{NN}(\bu, t)\right)\dt \p_{x^{i}}u(x, t) \\
        & - \sum_{i = 0}^d\sum_{j=0}^{d-i-1} (-1)^{j+1} \p_{x^{j}} \left(\mt(x, t+\tau)\p_{\p_{x^{j+i+1}} u(x, t)} \mathcal{D}_{NN}(\bu, t)\right)\dt \p_{x^{i}}u(x, t) \;, \\
        & \hspace{0.6\textwidth} x \in \partial \Omega, \; t \in [t, T) \,.
    \end{split}
\end{equation}
Note that adjoint PDE needs to be solved backward in time, and one would require access to $u(x, t), \forall x \in \Omega, \; 0 \leq t \leq T$. After solving for the Lagrangian variables, $\lambda(x, t)$ and $\mu(x, t)$, we can compute the required gradients as follows:
\begin{equation}
    \begin{split}
        \dt\mathcal{L} &= - \iw \mt(x, t) \pth \mathcal{D}_{NN}(\bu, t; \theta) dxdt + \iw \mt(x, t) \pth \mathcal{D}_{NN}(\bu, t-\tau; \theta) dxdt \\
        & - \int_{\Omega} \mt(x, 0) \int_{-\tau}^0 \pth \mathcal{D}_{NN}(h(x, t), \p_x h(x, t), \p_{xx} h(x, t), x, t; \theta)dt dx  \,.
    \end{split}
\end{equation}
If we restart the above derivation by taking derivative of the Lagrangian (equation \ref{eq(SI): lagrangian}) w.r.t. $\phi$, we will arrive at the same adjoint PDEs (equations \ref{eq(SI): adjoint PDEs} \& \ref{eq(SI): adjoint PDEs BCs}), and the required gradient will be given by,
\begin{equation}
    \begin{split}
        d_{\phi}\mathcal{L} &= - \iw \lt(x, t) \pp \mathcal{F}_{NN}(\bu, t; \phi) dxdt  \,.
    \end{split}
\end{equation}
Finally, using any stochastic gradient descent algorithm, we can find the optimal values of the weights $\phi$ and $\theta$.

\section{Experimental Setup}
\label{sec: gnCM Experimental Setup}
%auto-ignore
\subsection{Architectures}
The architectures used to generate the results corresponding to different experiments are provided in table \ref{table: All architecture}. The implementation details of the various biological and carbonate constraints imposed on the neural closure terms in experiments 2a \& 2b are also provided.

\subsection{Hyperparameters}
\label{sec: Hyperparameters}
The values of the tuned training hyperparameters corresponding to different experiments are listed next. In all the experiments, the number of iterations per epoch is calculated by dividing the number of time-steps in the training period by the batch-size multiplied by the length of short time-sequences, adding 1, and rounding up to the next integer.

\subsubsection{Experiments-1a:}
For training, we randomly select short time-sequences spanning 3 time-steps (batch-time) and extract data at every time-step to form batches of size 16; 4 iterations per epoch are used; an exponentially decaying learning rate (LR) schedule is used with an initial LR of 0.075, the decay rate of 0.97, and 4 decay steps; the RMSprop optimizer is employed; training is for a total of 150 epochs. $\mathcal{L}_1$ and $\mathcal{L}_2$ regularization with factors of $1.5\times 10^{-3}$ and $1\times 10^{-5}$, respectively, is used; the weights are pruned if the value drops below $5\times 10^{-3}$.

\subsubsection{Experiments-1b:}
\textbf{Only Markovian closure case:} We randomly select short time-sequences spanning 30 time-steps (batch-time) and  extract data at every other time-step to form batches of size 2. In total 24 iterations per epoch are used, with every 8 of them belonging to one of the $(N_x, Re)$ pairs. An exponentially decaying learning rate (LR) schedule with an initial LR of 0.025, a decay rate of 0.95, and 24 decay steps are used; the RMSprop optimizer is used; we train for a total of 30 epochs. We also use both $\mathcal{L}_1$ and $\mathcal{L}_2$ regularization for the weights of the neural network with factors of $5 \times 10^{-4}$ and $5\times 10^{-4}$, respectively, along with pruning of the weights if their value drops below $5 \times 10^{-3}$.

\noindent\textbf{Both Markovian and non-Markovian closures case:} A batch-time of 30 time-steps is used with data extracted at every other time-step to form batches of size 2; 32 iterations per epoch are used, with every 8 of them belonging to one of the $(N_x, Re)$ pairs; an exponentially decaying learning rate (LR) schedule is employed with an initial LR of 0.01, the decay rate of 0.95, and 32 decay steps; the RMSprop optimizer is used; we train for a total of 30 epochs. We also use both $\mathcal{L}_1$ and $\mathcal{L}_2$ regularization for the weights of the neural network with factors of $1.5 \times 10^{-3}$ and $1\times 10^{-5}$, respectively, along with pruning of the weights if their value drops below $5 \times 10^{-3}$.

\subsubsection{Experiments-2a:}
Parameter values used in the ocean acidification model are (adopted from \cite{tian2015model, eknes2002ensemble}): $g_{max} = 0.7~day^{-1}$; $k_W = 0.08~m^{-1}$; $K_N = 0.5~mmol~N~m^{-3}$; $K_P = 0.25~mmol~N~m^{-3}$; $m_P = 0.08~day^{-1}(mmol~N~m^{-3})^{-1}$; $m_Z = 0.030~day^{-1}(mmol~N~m^{-3})^{-1}$; $\mu_{max} = 2.808~day^{-1}$; $ \alpha = 0.14~(W~m^{-2}~day)^{-1}$; $ \beta = 0.0028~(W~m^{-2}~day)^{-1}$; $\epsilon = 0.015~day^{-1}$; $\lambda = 0.3$; $\gamma = 0.4$; a sinusoidal variation in $I_o(t)$; linear vertical variation in total biomass $T_{bio}(z)$ from $10~mmol~N~m^{-3}$ at the surface to $20~mmol~N~m^{-3}$ at $z = 100~m$; $K_{z_b} = 0.0864~(m^2/day)$; $K_{z_0} = 8.64~(m^2/day)$; $\gamma_t = 0.1~m^{-1}$; $D_z = -100~m$; and $\rho_w = 1000~kg/m^3$. For training, we randomly select short time-sequences spanning 3 time-steps (batch-time) and  extract data at every other time-step to form batches of size 4; we use 26 iterations per epoch;  an exponentially decaying learning rate (LR) schedule is used with an initial LR of 0.075, the decay rate of 0.97,  and 26 decay steps; the RMSprop optimizer is adopted; training is terminated at 200 epochs. We also use both $\mathcal{L}_1$ and $\mathcal{L}_2$ regularization for the weights of the neural network with factors of $1.5 \times 10^{-3}$ and $1\times 10^{-3}$, respectively, along with pruning of the weights if their value drops below $5 \times 10^{-3}$. 

\subsubsection{Experiments-2b:}
We use a batch-time of 3 time-steps with data extracted at every other time-step to form batches of size 8; we use 26 iterations per epoch; an exponentially decaying learning rate (LR) schedule is applied with an initial LR of 0.075, the decay rate of 0.97,  and 26 decay steps; the RMSprop optimizer is employed; training is terminated at 200 epochs. We also use both $\mathcal{L}_1$ and $\mathcal{L}_2$ regularization for the weights of the Markovian closure with factors of $1.5 \times 10^{-3}$ and $1\times 10^{-3}$, respectively, along with pruning of the weights if their value drops below $5 \times 10^{-3}$. For the neural network in the non-Markovian closure term, only $\mathcal{L}_2$ regularization with a factor of $1\times 10^{-5}$ is used.

Finally, for all the experiments and their multiple repeats with the exact same tuned hyperparameters, we provide variation of training and validation error as training progresses (figure \ref{fig: gnCM Loss plots}).

%auto-ignore
\begin{table}
\resizebox{1\textwidth}{!}{%
\caption{
Architectures for different generalized neural closure models used in the four sets of experiments. We explicitly provide the constraints on the weights and output layer of neural networks used in different experiments. $\{w_i\}_{i=1}^4$ are row vectors of the weight matrix. ``Effective'' number of trainable weights does not count the ones which are not free or are overwritten due to the imposed constraints. $C_P$, $C_Z$, and $C_D$ are the carbon-nitrogen ratios for phytoplanktons, zooplanktons, and detritus, respectively. $\rho_w$ is seawater density.}
\label{table: All architecture}
\begin{tabular}{|p{0.1\textwidth}|p{0.2\textwidth}|p{0.075\textwidth}|p{0.375\textwidth}|p{0.2\textwidth}|p{0.05\textwidth}|p{0.08\textwidth}|p{0.125\textwidth}|}
\hline
\multirow{3}{*}{\parbox{0.1\textwidth}{\textbf{Experi-ments}}}      &       \multicolumn{3}{c|}{\textbf{Markovian Term}}     &            \multicolumn{4}{c|}{\textbf{Non-Markovian Term}}       \\
\cline{2-4}     \cline{5-8}
&       \multirow{2}{*}{\textbf{Architecture}}      &       \multirow{2}{*}{\textbf{Act.}}        &       \multirow{2}{*}{\parbox{0.3\textwidth}{\textbf{Trainable Weights}}}       &       \multirow{2}{*}{\textbf{Architecture}}      &       \multirow{2}{*}{\textbf{Act.}}      &       \multirow{2}{*}{\textbf{Delays}}        &       \multirow{2}{*}{\parbox{0.125\textwidth}{\textbf{Trainable Weights}}}      \\
&       &       &       &       &       &       &       \\
\hline
\multirow{3}{*}{\parbox{0.15\textwidth}{1a}}      &       \multicolumn{2}{c|}{$\mathcal{F}_{NN}$}        &       \multirow{3}{*}{4}     &       \multicolumn{4}{c|}{}        \\
\cline{2-3}    
&       Input layer with 4 neurons      &       none        &       &       \multicolumn{4}{c|}{}       \\
\cline{2-3}     
&       Dense output layer with 1 neurons        &       linear       &       &       \multicolumn{4}{c|}{}       \\
\hline
\multirow{8}{*}{\parbox{0.15\textwidth}{1b}}      &       \multicolumn{2}{c|}{$\mathcal{F}_{NN}$}        &       \multirow{8}{*}{4}     &       \multicolumn{2}{c|}{$\mathcal{G}_{NN}$}        &       \multirow{8}{*}{0.075}       &       \multirow{8}{*}{198}        \\
\cline{2-3}     \cline{5-6}
&       Input layer with 4 neurons      &       none        &       &       Input layer with 5 neurons      &       none        &       &       \\
\cline{2-3}     \cline{5-6}
&       Dense output layer with 1 neurons        &       linear       &       &       Dense hidden layer with 10 neurons
        &       swish        &       &       \\
     \cline{5-6}
&       &       &       &       Dense hidden layer with 7 neurons
      &       swish      &       &       \\
\cline{5-6}
&      &      &       &       2 Dense hidden layers with 5 neurons
      &       swish      &       &       \\
\cline{5-6}
&      &      &       &       Dense hidden layer with 3 neurons
      &       swish      &       &       \\
\cline{5-6}
&      &      &       &       Dense output layer with 1 neuron
      &       linear      &       &       \\
\cline{5-6}
&      &      &       &       \multicolumn{2}{c|}{Multiply output with $|u|$
}      &       &       \\
\hline
\multirow{5}{*}{\parbox{0.15\textwidth}{2a}}       &       \multicolumn{2}{c|}{$\mathcal{F}_{NN}$}        &       \multirow{5}{*}{\parbox{0.4\textwidth}{18 (effective) \newline $w = \begin{bmatrix} -(w_2 + w_3 + w_4) \\ w_2 \\ w_3 \\ w_4 \\ -C_P w_2 -C_Z w_3 - C_D w_4 \\ (w_2 + w_3 + w_4) / \rho_w \end{bmatrix}$} }     &       \multicolumn{4}{c|}{}        \\
\cline{2-3}    
&       Input layer with 6 neurons      &       none        &       &       \multicolumn{4}{c|}{}       \\
\cline{2-3}     
&       Dense output layer with 6 neurons        &       linear       &       &       \multicolumn{4}{c|}{}       \\
&        &       &       &       \multicolumn{4}{c|}{}       \\
&        &       &       &       \multicolumn{4}{c|}{}       \\
\hline
\multirow{5}{*}{\parbox{0.15\textwidth}{2b}}      &       \multicolumn{2}{c|}{$\mathcal{F}_{NN}$}        &       \multirow{5}{*}{\parbox{0.2\textwidth}{4 (effective) \newline $w = \begin{bmatrix} -w_3 \\ 0 \\ w_3 \\ -C_Z w_3 \\ w_3 / \rho_w \end{bmatrix}$} }     &       \multicolumn{2}{c|}{$\mathcal{G}_{NN}$}        &       \multirow{5}{*}{2.5}       &       \multirow{5}{*}{65}        \\
\cline{2-3}     \cline{5-6}
&       Input layer with 4 neurons      &       none        &       &       Input layer with 4 neurons      &       none        &       &       \\
\cline{2-3}     \cline{5-6}
&       Dense output layer with 5 neurons        &       linear       &       &       2 Dense hidden layer with 5 neurons
        &       swish        &       &       \\
     \cline{5-6}
&       &       &       &       Dense output layer with 4 neurons
      &       linear      &       &       \\
     \cline{5-6}
&       &       &       &       \multicolumn{2}{c|}{$\mathcal{G}_{NN} = \begin{bmatrix} -(\mathcal{G}_{NN}^1 + \mathcal{G}_{NN}^2) \\ \mathcal{G}_{NN}^1 \\ \mathcal{G}_{NN}^2 \\ \mathcal{G}_{NN}^3 \\ \mathcal{G}_{NN}^4 / \rho_w \end{bmatrix}$}      &       &       \\
\hline
\end{tabular}}
\end{table}

\begin{figure}[]
	\centering
	\subfloat[][Experiments-1a]{\includegraphics[width=0.85\textwidth]{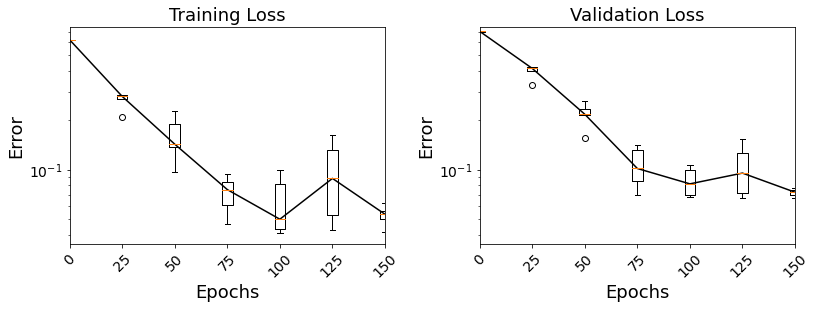}} \\
	\subfloat[][Experiments-1b (\textit{g}nCM with only Markovian closure term)]{\includegraphics[width=0.85\textwidth]{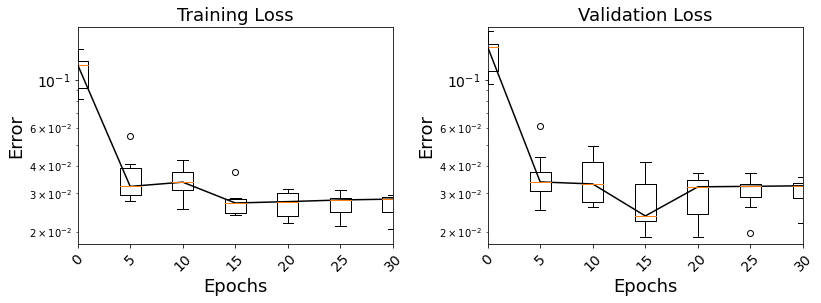}} \\
	\subfloat[][Experiments-1b (\textit{g}nCM with both Markovian and non-Markovian closure terms)]{\includegraphics[width=0.85\textwidth]{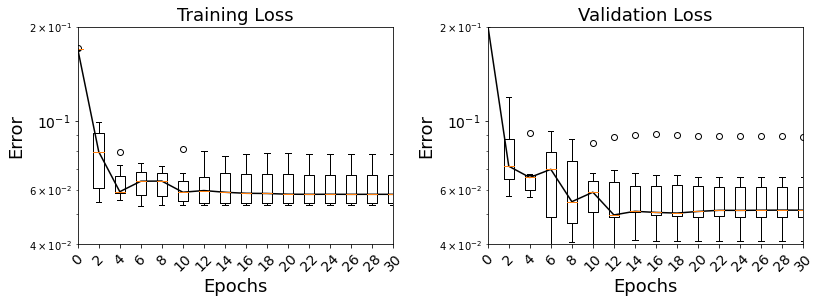}} 
	\caption{Variation of training (left column) and validation (right column) loss with epochs, for each of the experiments-1a, 1b, 2a, and 2b. We use boxplots to provide statistical summaries for multiple training repeats done for each set of experiments. The box and its whiskers provide a five-number summary: minimum, first quartile (Q1), median (orange solid line), third quartile (Q3), and maximum, along with outliers (black circles) if any. These results accompany the architectures detailed in table \ref{table: All architecture}. \emph{(cont.)}}
	\label{fig: gnCM Loss plots}
\end{figure}
\begin{figure}[]
\ContinuedFloat
	\centering
	\subfloat[][Experiments-2a]{\includegraphics[width=0.85\textwidth]{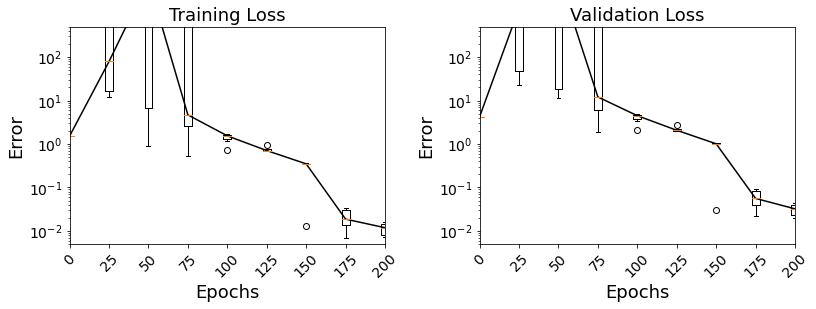}} \\
	\subfloat[][Experiments-2b]{\includegraphics[width=0.85\textwidth]{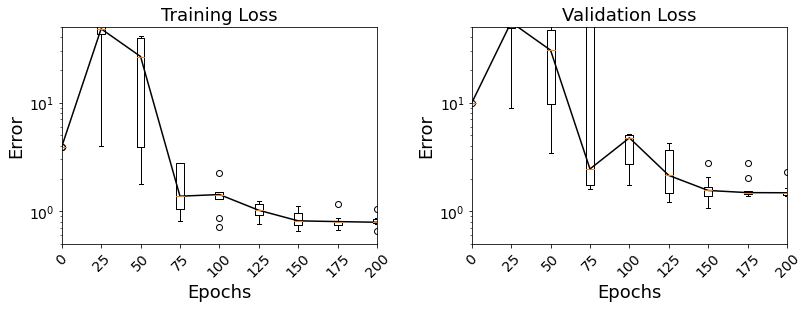}}
	\caption{Variation of training (left column) and validation (right column) loss with epochs, for each of the experiments-1a, 1b, 2a, and 2b. We use boxplots to provide statistical summaries for multiple training repeats done for each set of experiments. The box and its whiskers provide a five-number summary: minimum, first quartile (Q1), median (orange solid line), third quartile (Q3), and maximum, along with outliers (black circles) if any. These results accompany the architectures detailed in table \ref{table: All architecture}.}
\end{figure}

\bibliographystyle{unsrtnat}
\bibliography{mseas,ml}
\makeatletter\@input{suppxx.tex}\makeatother